\documentclass{article}

\usepackage{arxiv}

\usepackage[utf8]{inputenc} 
\usepackage[T1]{fontenc}    
\usepackage{hyperref}       
\usepackage{url}            
\usepackage{booktabs}       
\usepackage{amsfonts}       
\usepackage{nicefrac}       
\usepackage{microtype}      
\usepackage{lipsum}		
\usepackage{graphicx}
\usepackage{natbib}
\usepackage{doi}

\title{Training-Free Action Recognition and Goal Inference with Dynamic Frame Selection}

\author{
        {\hspace{1mm} Yeo Keat Ee}\\
        Centre for Frontier AI Research\\
        Agency for Science, Technology\\
        and Research (A*STAR)\\
        Singapore\\
	\texttt{ee\_yeo\_keat@cfar.a-star.edu.sg}\\
	\And
	{\hspace{1mm}Hao Zhang}\\
	Institute of High Performance Computing\\
        Agency for Science, Technology\\
        and Research (A*STAR)\\
        Singapore \\
	\texttt{zhang\_hao@ihpc.a-star.edu.sg} \\
	\And
	{\hspace{1mm}Alexander Matyasko}\\
        Centre for Frontier AI Research\\
        Agency for Science, Technology\\
        and Research (A*STAR)\\
        Singapore\\
	\texttt{alexander\_matyasko@cfar.a-star.edu.sg} \\
	\And
	{\hspace{1mm}Basura Fernando}\\
        Centre for Frontier AI Research\\
        Agency for Science, Technology\\
        and Research (A*STAR)\\
        Singapore\\
	\texttt{fernando\_basura@cfar.a-star.edu.sg} \\
}

\date{}

\renewcommand{\undertitle}{}

\usepackage{amsmath}
\usepackage{multirow}
\usepackage{graphicx}
\usepackage{booktabs}
\usepackage{cleveref}

\usepackage{xcolor}

\begin{document}
\maketitle

\begin{abstract}
We introduce VidTFS, a \textit{\textbf{T}raining-free}, \textit{open-vocabulary} video goal and action inference framework that combines the frozen vision foundational model (VFM) and large language model (LLM) with a novel dynamic \textit{\textbf{F}rame \textbf{S}election} module.
Our experiments demonstrate that the proposed frame selection module improves the performance of the framework significantly. 
We validate the performance of the proposed VidTFS on four widely used video datasets, including CrossTask, COIN, UCF101, and ActivityNet, covering goal inference and action recognition tasks under open-vocabulary settings without requiring any training or fine-tuning.  
The results show that VidTFS outperforms pre-trained and instruction-tuned multimodal language models that directly stack LLM and VFM for downstream video inference tasks. 
Our VidTFS with its adaptability shows the future potential for generalizing to new training-free video inference tasks.

\end{abstract}    
\section{Introduction}\label{sec:intro}

Video understanding tasks such as action recognition and anticipation have significantly progressed due to the scaling up of video data \cite{Kay2017TheKH,activitynet,Damen2020RescalingEV} and the development of powerful foundational models \cite{clip, bain2021frozen,wang2022internvideo}. 
As large language models (LLMs) and vision foundational models (VFMs) continue to evolve, many works have leveraged them to perform various vision tasks with few or no examples and without additional training. 
However, foundational model development~\cite{clip,li2022blip} and instruction tuning~\cite{liu2024visual,lin2023video, zhang2023video} requires large-scale datasets and computations which is not practical for every downstream task.
This inspires a new direction of \textit{training-free}, \textit{open-vocabulary} vision-language understanding~\cite{xu2022simple, udandarao2023sus}. 

Specifically, LLMs trained on the large-scale corpus emerge with open-vocabulary capability and able to be generalized to unseen tasks \cite{brown2020language,Touvron2023LLaMAOA}. 
Recent multimodal language model (MLM) that incorporates LLM with VFM (e.g., LLaMA+CLIP) \cite{liu2024visual, Ye2023mPLUGOwlME} shows strong zero-shot ability on several downstream visual tasks, e.g. classification, detection, segmentation. 
The general problem-solving ability of LLMs and MLMs points to the new ways of solving downstream video inference tasks without requires fine-tuning and under open-vocabulary setting. 
However, most of the existing works focus on processing static images \cite{udandarao2023sus,novack2023chils,xu2022simple}, whereas research on video inference with training-free and open-vocabulary setting still needs to be explored.
Our motivation for developing a video inference framework that is both training-free and capable of open vocabulary inference stems from three key aspects:
First, the training-free condition is desired. 
Fully fine-tuning a video VFM to process long videos demands high computational resources and usually not ideal for downstream tasks.
Second, frame selection is necessary because untrimmed video naturally contains non-relevant information which not only burdened video processing but also affect model performance.
Finally, vanilla VFM (e.g., CLIP and BLIP) learns from image-text pair and doesn't exhibit generative deduction ability as LLM; thereby, we leverage the LLM on top of VFM for open-vocabulary inference, similar spirit as generic multimodal language models (e.g., LLaVA \cite{liu2024visual}).

\begin{figure}[t]
  \centering
   \includegraphics[width=0.9\linewidth]{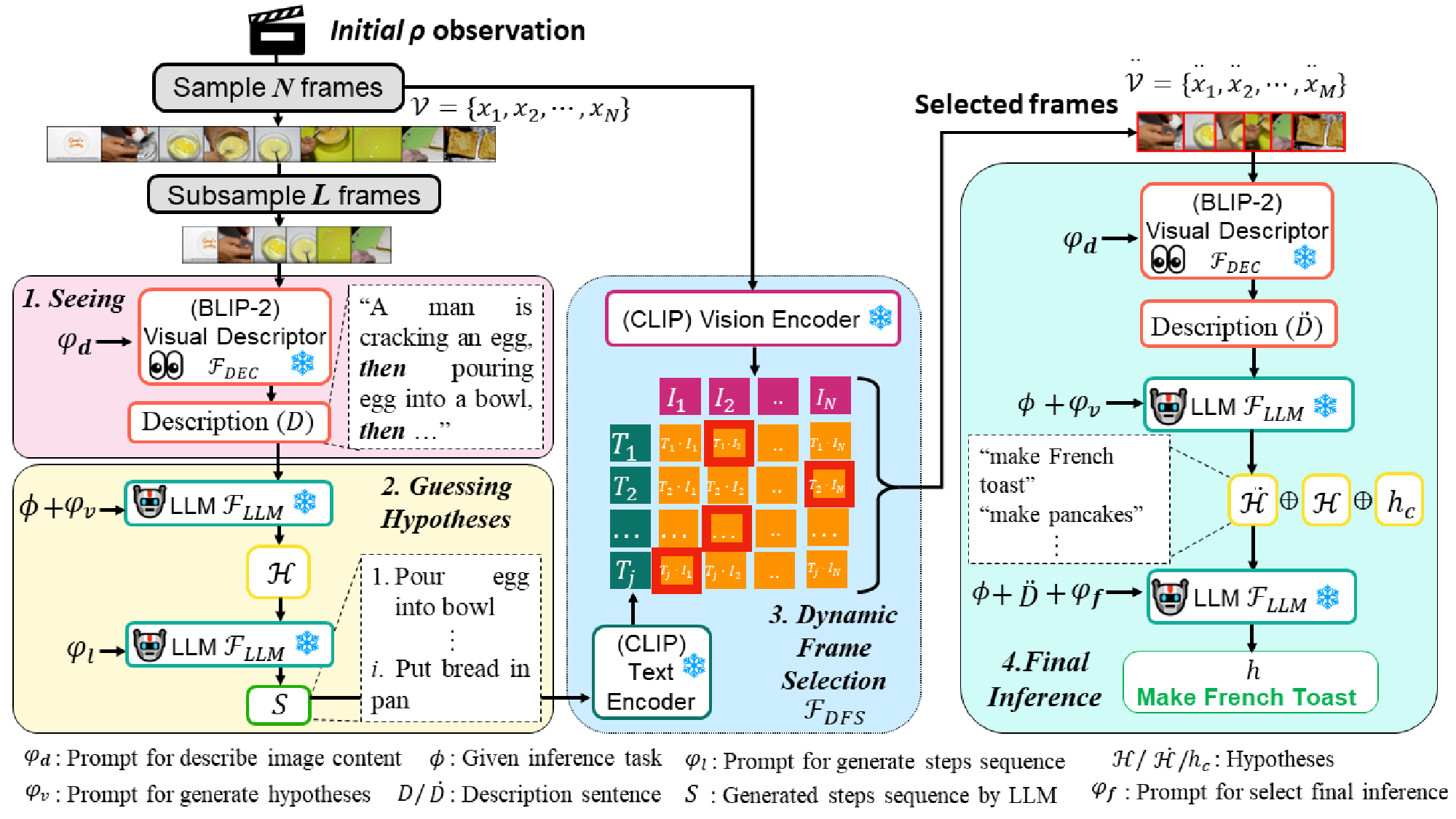}
   \caption{VidTFS contains four stages: \textit{See}, \textit{Guess}, \textit{Select}, and \textit{Infer}. (1). Seeing through Visual Descriptor (i.e., BLIP-2) translates visual frames into dense textual descriptions. (2). Guessing by LLM generate hypotheses ($\mathcal{H}$) and corresponding sub-events (steps). (3). Selecting frames using CLIP reduce irrelevant frames. (4). Inferring final answer by using selected frames with the ``see'' \& ``guess'' process again. Best viewed on computer full screen.}
   \vspace{-1em}
   \label{fig:fullframework}
\end{figure}

To achieve these objectives, we proposed the VidTFS as shown in Figure \ref{fig:fullframework}, a training-free framework that is capable of open-vocabulary video goal inferences and action recognition. VidTFS leverages frozen VFM (BLIP-2~\cite{li2023blip} and CLIP~\cite{clip}) and LLM (Vicuna \cite{zheng2023judging}), without specific tuning on downstream tasks. Specifically, we rely on a visual descriptor BLIP-2~\cite{li2023blip} model to describe what it sees. Then, we propose a dynamic frame selection module (a.k.a. the evidence selector) by using the frozen CLIP model~\cite{clip} and the LLM-generated script (program) of the hypotheses to select evidence frames. Finally, we rely on the excellent reasoning capabilities of LLM to make inference based on the selected evidence frames and corresponding descriptions.
We evaluate the VidTFS across several video datasets on open-vocabulary inferences, covering goal inference and action recognition tasks. 
Experimental results show that our VidTFS achieve better performance under training-free and open-vocabulary settings.
We observe that the VFMs excel in describing visual content but need help with reasoning. Meanwhile, by introducing the LLMs reasoning capabilities, we find our VidTFS can handle video tasks that demand comprehension without training or fine-tuning.
In summary, our contributions are as follows:

\noindent
\textbf{Training-Free VidTFS}: We introduce VidTFS, which composes VFM + LLM for open-vocabulary video inference tasks without requiring training. 
In VidTFS, VFM and LLM fulfil distinct roles such as ``eye'', ``guesser'', and ``selector'', and they exchange information through language while analyzing videos.
   
\noindent   
\textbf{Evidence Selector}: We propose a training-free dynamic frame selection module to identify relevant frames to support the video inference tasks. 
This process involves using an LLM to generate hypotheses and the scripts, then matching each with visual frames by using frozen VFM (e.g. CLIP) to select the relevant frames as support evidence.

\noindent   
\textbf{Generalization for Different Video Inference Task}: 
We evaluated VidTFS on four video datasets, covering tasks like goal inference and action recognition. 
The results show that VidTFS either outperforms or matches the performance of the SOTA multimodal LLM, showing its generalizability and robustness.

\section{Related Work}
\label{sec:relatedwork}

\noindent

\textbf{Supervised learning for video understanding} has been extensively studied in the era of foundational models. With the success of foundational models on static images (e.g., CLIP \cite{clip}), numerous video models have been proposed to learn visual video representations from large-scale data. Representative works included \cite{wang2022internvideo, vificlip, xclip, yang2023aim, wasim2023vita, sun2019videobert}. 
ViFi-CLIP \cite{vificlip} shows that fine-tuning CLIP with large-scale video data leads to better video classification. 
Whereas, with less data, prompt tuning CLIP can help reduce the risk of overfitting. 
Similarly, Vita-CLIP \cite{wasim2023vita} proposes learnable prompts at different temporal levels to align video-text pairs. 
While \cite{ju2022prompting} add learnable prompt vectors to the CLIP text encoder to create action classifiers.
AIM \cite{yang2023aim} plugs adapters into backbones to reduce training computations and alleviate overfitting. 
These methods require supervised training with substantial video annotation data. 
In contrast, the proposed VidTFS is applied to video recognition tasks without any training that enhances adaptability of the foundation models.

\textbf{Instructional tuning of videos} uses both large language models and vision foundational models~\cite{zhang2023video, lin2023video, videochatgpt, Ye2023mPLUGOwlME, Zhao2023AntGPTCL}. 
These models are adapted using large-scale VQA datasets.
They show robust zero-shot and open-vocabulary generation capability on unseen downstream video tasks. 
Specifically, Video-LLaMA \cite{zhang2023video} uses frozen VFM (ViT \cite{dosovitskiy2020image}) and LLM (e.g., Vicuna \cite{vicuna2023}, LLaMA), and only learn the Q-Fromer \cite{li2023blip}.
Similarly, Video-LLaVA \cite{lin2023video} combines LanguageBind \cite{zhu2023languagebind} and Vicuna for video encoding and language processing, and includes a projection layer to link visual and text tokens together. 
VideoChat \cite{videochatgpt} uses two separate VFMs to create visual captions and visual embeddings. 
These are then combined and fed into a LLM for question and answer processing. 
The mPLUG-Owl~\cite{Ye2023mPLUGOwlME} model adopts a cross-attention mechanism with learnable queries to project visual tokens into textual space. 
VidTFS differs by not needing to fine-tune (e.g. Q-Former or linear projection), leveraging the training-free interaction between LLM and VFMs via language and dynamic frame selection for video inferences.
\noindent

\textbf{Training-free open-vocabulary image understanding} gaining extensive research interests by treating large-scale pre-trained models as tools. 
Many studies, like \cite{menon2022visual,novack2023chils}, utilize strong zero-shot capabilities of pre-trained CLIP and combine it with ChatGPT-3.5 for open-vocabulary image classification. Other research efforts focus on solely enhancing CLIP's ability to understand different vocabularies without additional training, as in \cite{udandarao2023sus,xu2022simple}. Specifically, VisDesc \cite{menon2022visual} expands unseen categories using detailed text descriptions by inquiring ChatGPT and then pairs images with these descriptions using a frozen CLIP model. Similarly, the CHiLS \cite{novack2023chils} replaces coarse-defined categories with more specific sub-categories. These sub-categories are created using label hierarchies or consulting ChatGPT and then matched with visual content using CLIP. Besides, SuS-X \cite{udandarao2023sus} creates a support set that includes open categories by stable-diffusion \cite{rombach2022high} or retrieval methods. Using CLIP models, it then measures the distance between a query image and the support set, broadcasting labels from the support set to the query. Xu et.al \cite{xu2022simple} utilize off-the-shelf mask generators and frozen CLIP for open-vocabulary semantic segmentation. VidTFS also employs ready-to-use BLIP-2, CLIP, and Vicuna, but it differs in handling dynamic video inputs and introduces frame selection for narrowing down evidence using foundational models.




\textbf{Training-free open-vocabulary video understanding} also makes use of pre-trained foundational models' perception and reasoning abilities to tackle new video tasks. Example works like \cite{Zeng2022SocraticMC, chen2023video} involve using several large pre-trained models as tools. These models function in roles of perception and reasoning and interact with each other through language. Specifically, the Socratic Models \cite{Zeng2022SocraticMC} introduce a technique of multimodal prompting across multiple models. This involves a combination of a vision-language model (like CLIP with BERT/GPT), an Audio Large Model \cite{bapna2022mslam}, and a Large Language Model (LLM). This approach exchanges information between these large models through text and can handle new video tasks. Similarly, VideoChatCaptioner \cite{chen2023video} set up a conversation between ChatGPT and BLIP-2, with ChatGPT asking questions and BLIP-2 answers based on the input video. The video's description is progressively enhanced through multiple rounds of automated conversation. Our VidTFS also aligns with this direction, focusing on interactions between LLMs and VFMs. Our main difference from existing methods is that we have found that focusing on the most important parts of a video using an evidence selector improves open-vocabulary video inference performance with lesser computations.

\section{Method}

Our VidTFS framework $\mathcal{F}$ solve the open-vocabulary inference task $\phi$ (e.g., ``goal inference'') by processing natural video $\mathcal{V} = \{v_1, v_2, \ldots, v_N\}$, which consists of $N$ uniformly sampled frames. 
We infer the most likely hypothesis $h$ based on the video observation without training or fine-tuning. 
\begin{equation}
    {h} = \mathcal{F}(\mathcal{V},\phi)
\end{equation}
Examples of the hypothesis include \textit{make French toast} for inferring goals in cooking videos, and \textit{baby crawling} for recognising actions in videos.
An overview of VidTFS modular framework is shown in Figure ~\ref{fig:fullframework}, it uses three frozen foundational models: BLIP-2 as the visual descriptor $\mathcal{F}_{\text{DEC}}$, Vicuna as the LLM agent $\mathcal{F}_{\text{LLM}}$, and CLIP as the dynamic frame selector $\mathcal{F}_{\text{DFS}}$. Given target task $\phi$ and video $\mathcal{V}$, these models work together in four stages: \textit{See}, \textit{Guess}, \textit{Select}, and \textit{Infer} . Details of each stage are as follows. 

\paragraph{\textbf{Seeing through Visual Descriptor}}: 
We further uniformly sub-sample $L$ out of $N$ selected frames.
The visual descriptor $\mathcal{F}_{\text{DEC}}$ takes each sampled frame $x_i$ as input and outputs a caption (text description) sentence $c_i$.
We use BLIP-2~\cite{li2023blip} (FLanT5-XXL) model as a visual descriptor and use a prompt $\varphi_d$ to obtain the frame description, for example, $\varphi_d=$\emph{what is the content of the image?}.
The sequence of all frame captions is denoted by $\mathcal{C}=\{c_1, c_2, \cdots, c_L\}$ and there are a total of $L$ captions. 
Next, we concatenate the captions in $\mathcal{C}$ into a single continuous description paragraph $\mathcal{D}$ using the word ``\textit{then}'' to link them up so that $\mathcal{D}$ follows the form of ``\texttt{<caption 1>, then, <catpion 2>, then, ... <caption L>}''.
\paragraph{\textbf{Guessing Hypotheses with LLM}}: 
We use a LLM ($\mathcal{F}_{\text{LLM}}$) to guess the top-$k$ initial hypotheses, $\mathcal{H}=\{h_1, h_2,\cdots, h_k\}$  for the given inference task, $\phi$ (\cref{eq:guess_1}), with an instructional prompt $\varphi_v$. 
Here $\varphi_v$ is \texttt{I want to perform <task>, generate top-<$k$> hypotheses, given <text>}.  
\begin{equation}
\mathcal{H} = \mathcal{F}_{\text{LLM}}\bigr( \mathcal{D}, \varphi_v)
\label{eq:guess_1}
\end{equation}
Hereby, \texttt{<task>} is the task definition name (e.g. $\phi$ = goal inference) and \texttt{<text>} is description paragraph $\mathcal{D}$. Notably, we only show a simplified prompt version for quick reference and put the full instructional prompt in the supplementary section. We employ Vicuna \cite{zheng2023judging} as the $\mathcal{F}_{\text{LLM}}$. An example of guessed hypotheses $\mathcal{H}$=[``\textit{make French toast}'',``\textit{make pancakes}'', $\cdots$] --see also Figure \ref{fig:fullframework}.
We further expand each candidate hypothesis in $\mathcal{H}$ into a sequence of detailed events or steps, $\mathcal{S}$. 
We achieve this by using prompt $\varphi_l$ in the form of ``\texttt{List the steps to perform <hypotheses>}''.
\begin{equation}
\mathcal{S} = \mathcal{F}_{\text{LLM}}\bigr(\mathcal{H}, \varphi_l \bigr)
\label{eq:step_1}
\end{equation}
 Since there are $k$ potential hypotheses, we eventually have $k$ number of different step sequences. We gather all these sequences into $\mathcal{S} = \{[s^{h_1}_1, \cdots, ], \cdots,$ $[s^{h_k}_1, \cdots, s^{h_k}_i]\}$, re-flatten it into $\mathcal{S} = \{s_1, s_2,\cdots, s_j\}$ of $j$ total steps.
The reasons for expanding from $\mathcal{H}$$\rightarrow$$\mathcal{S}$ lies in two aspects. 
Firstly, steps contain more fine-grained information than the hypothesis, as a hypothesis is the outcome of executing a script containing a list of steps \cite{schank1975scripts}. 
A specific step often corresponds directly to visual details, whereas a hypothesis may lack visual representation. Conversely, video inference tasks like goal inference encompass multiple sub-steps essential for inference based on deductive reasoning.
By aligning the relevant frames with corresponding steps in a hypothesis, we can deduce that the hypothesis is a correct answer from the candidate set $\mathcal{H}$.

\paragraph{\textbf{Dynamic Frame Selection (DFS) by Evidence Selector}}: The evidence selector dynamically pick $M$ out of $N$ frames, creating a subset of frames $\mathcal{\ddot{V}}$ where $\mathcal{\ddot{V}} \subset \mathcal{V}$ that are relevant to the inference task. 
DFS mechanism finds the most relevant frame $\ddot{x}_i$ ( $ \ddot{x}_i \in \mathcal{\ddot{V}}$) for each hypothesized step $s_i$ in $\mathcal{S}$. 
We use frozen CLIP \cite{clip}, a two-tower vision-language encoder to implement $\mathcal{F}_{\text{DFS}}$.
\begin{equation}
\mathcal{\ddot{V}} = \mathcal{F}_{\text{DFS}}\bigr( \mathcal{V}, \mathcal{S}\bigr) = \{\ddot{x}_1, \ddot{x}_2, \cdots, \ddot{x}_M\}~~~~~~~~s.t.~M < N
\label{eq:select_1}
\end{equation}

Specifically, we use the CLIP vision encoder to extract features for all $N$ visual frames and the CLIP text encoder to process all $S$ steps in text form. We then calculate the cosine similarity between each $\langle step, frame \rangle$ pair as in Figure \ref{fig:fullframework} (middle). 
Afterwards, we select the top highest similarity score of $M$ frames resulting in a set of evidence frames $\mathcal{\ddot{V}}$.
We limit $M\le16$ to avoid picking out too many frames and post-process $\mathcal{\ddot{V}}$ to filter out duplicate frames.
With the evidence selector, we make sure that selected frames have diverse levels of information relevant to the task.

\paragraph{\textbf{Final Inference by LLM}}: We use the selected frames $\mathcal{\ddot{V}}$ to make inferences and generate the final hypothesis $h$ by LLM in an open-vocabulary manner. 
We repeat the process from ``Seeing through visual descriptor'' and ``Guessing hypothesis with LLM'' except that we do not require the LLM to generate the steps again. 
Instead, we infer a second set of top-$k$ hypotheses $\ddot{\mathcal{H}}$. 
Furthermore, we use the CLIP model to infer a single CLIP-based hypothesis using $\mathcal{\ddot{V}}$ and $\mathcal{H} \oplus \ddot{\mathcal{H}}$ which we denote as $h_c$. 
The $h_c$ is selected from the candidate hypotheses ($\mathcal{H}$ $\oplus$ $\ddot{\mathcal{H}}$) by finding the best-matched hypotheses to the mean-pooled visual features of those selected frames using CLIP visual and textual embeddings.
Then we take the hypothesis combination (operator denoted as $\oplus$) of all generated hypotheses, e.g., $\mathcal{H}$, $\ddot{\mathcal{H}}$ and $h_c$ as the candidates and let LLM infer the final hypothesis $h$ using the selected frame description $\ddot{\mathcal{D}}$ and final inference prompt $\varphi_f$ as follows:
\begin{equation}
h = {\mathcal{F}_{\text{LLM}}}_{ \{ \mathcal{H} \oplus \ddot{\mathcal{H}} \oplus h_c \} } ( \ddot{\mathcal{D}}, \varphi_f).
\label{eq:guess_final}    
\end{equation}
Here $\ddot{\mathcal{D}}$ is obtained from the BLIP-2 model after processing $\mathcal{\ddot{V}}$. As before, we use the term ``\textit{then}'' to form a coherent description of selected frames. 
The final inference prompt $\varphi_f$ follows the form of ``\texttt{I want to perform <task>, only select one answer from options <hypotheses>, given <text>}''. 
The full prompts format will be provided in supplementary.
Notably, we fill the \texttt{<hypotheses>} with $\mathcal{H} \oplus \ddot{\mathcal{H}} \oplus h_c$, and \texttt{<text>} with $\ddot{\mathcal{D}}$. 
For operator $\oplus$, we ablate choices of union operator $\cup$ and concatenation as shown in supplementary and choose the latter one. 


\section{Experiments}
\begin{figure}[tb]
\centering
\includegraphics[width=0.9\linewidth]{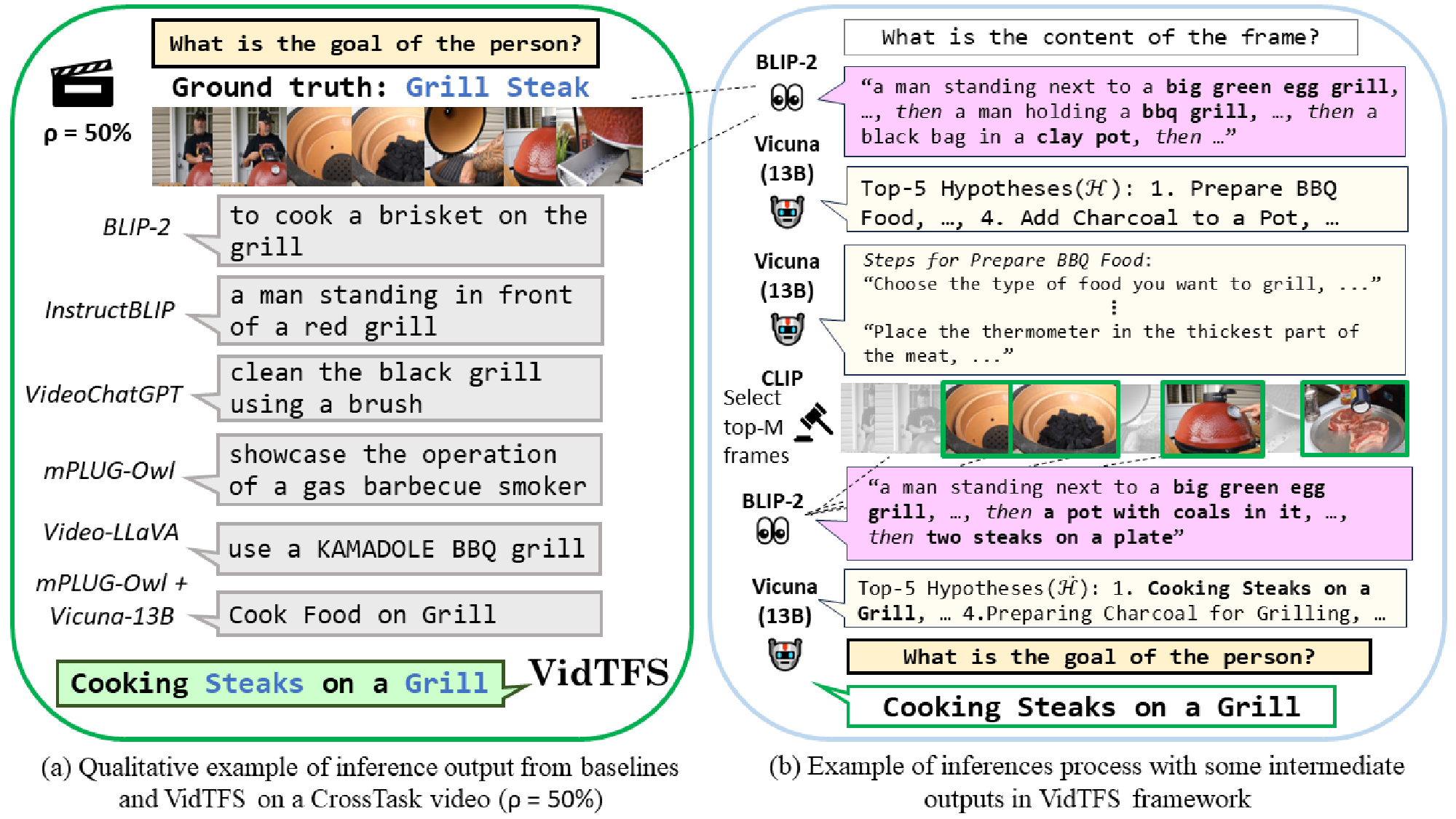}
\caption{Qualitative example of goal inference on CrossTask video. More qualitative examples are provided in supplementary.}
\label{fig:dl91S-X_VJs_sample1}
\end{figure}

\label{sec:experiments}
We evaluate the VidTFS on four relevant datasets, covering goal inference and action recognition tasks under training-free, open-vocabulary settings. 
We report the evaluation metrics, such as METEOR(M)~\cite{denkowski2014meteor}, SPICE(S)~\cite{anderson2016spice}, CIDEr(C)~\cite{vedantam2015cider}, that are commonly used in visual captioning literature \cite{anderson2018bottom}. 
To measure the semantic similarity between ground-truth answers and open-vocabulary inferences, we use BERTScore(B)~\cite{zhang2019bertscore}, SBERT(SB)~\cite{Reimers2019SentenceBERTSE}) as in \cite{sabir2023}.

\subsection{Datasets}
\noindent
\textbf{CrossTask} \cite{Zhukov2019} dataset consists of 4,700 instructional videos (avg. 5 minutes long) about 83 different daily tasks. We evaluate on the goal inference task by using the given validation set (360 untrimmed videos) that covered 18 primary tasks and only use the task labels as our ground truth goal labels during evaluation. 
\textbf{COIN} \cite{coin2019} dataset comprises 11,827 instructional videos (avg. 2.36 minutes long) with 180 distinct tasks. We evaluate the test set of 2,797 untrimmed videos and only use the corresponding task label as the goal label.
\textbf{UCF101} \cite{soomro2012ucf101} dataset is a widely utilized benchmark for action recognition tasks. It consists of 13,320 short videos (avg. 7.5 seconds long) and encompasses 101 distinct action classes, each video depicts a single action. We perform the action recognition evaluation on all three test splits of the dataset.
\textbf{ActivityNet-v1.3} \cite{activitynet} dataset contains 19,994 untrimmed YouTube videos (avg. 2 minutes long) consisting of 200 action classes. We evaluate using their validation set of 4,926 videos for the action recognition task.

\subsection{Goal Inference}
\label{sec:zsgoal}

\begin{table}[t]

\scriptsize
\centering
\resizebox{\textwidth}{!}{
\begin{tabular}{l|ccccc|ccccc|ccccc}
\toprule 
 \multirow{2}{*}{\textbf{CrossTask}} & 
 \multicolumn{5}{c|}{\begin{math}\rho\end{math} = 10\%} & 
 \multicolumn{5}{c|}{\begin{math}\rho\end{math} = 30\%} & 
 \multicolumn{5}{c}{\begin{math}\rho\end{math} = 50\%} \\
 
 & \multirow{1}{*}{\footnotesize{S}} & \multirow{1}{*}{\footnotesize{C}} 
 & \multirow{1}{*}{\footnotesize{M}} & \multirow{1}{*}{\footnotesize{B}} & \multirow{1}{*}{\footnotesize{SB}}   
 
 & \multirow{1}{*}{\footnotesize{S}} & \multirow{1}{*}{\footnotesize{C}} 
 & \multirow{1}{*}{\footnotesize{M}} & \multirow{1}{*}{\footnotesize{B}} & \multirow{1}{*}{\footnotesize{SB}} 
 
 & \multirow{1}{*}{\footnotesize{S}} & \multirow{1}{*}{\footnotesize{C}} 
 & \multirow{1}{*}{\footnotesize{M}} & \multirow{1}{*}{\footnotesize{B}} & \multirow{1}{*}{\footnotesize{SB}}\\

 \midrule
BLIP-2\cite{li2023blip} & 13.3 & 27.2 & \underline{11.6} & 15.9 & 32.2
& 11.7 & 24.2 & 11.6 & 16.7 & 33.1
& 12.6 & 24.8 & 12.2 & 17.5 & 34.5
\\

InstructBLIP\cite{instructBLIP} & 6.2 & 6.6 & 5.5 & -0.2 & 23.6 & 4.9 & 4.6 & 4.7 & -0.4 & 22.4
& 4.8 & 4.2 & 4.5 & -0.3 & 22.8
\\

Video-ChatGPT\cite{videochatgpt} & 9.0 & 14.9 & 10.5 & 11.9 &
35.4 & 10.0 & 18.1 & 12.1 & 15.2 &
38.4 & 9.7 & 23.1 & 12.5 & 16.6 & 39.6
\\

mPLUG-Owl \cite{Ye2023mPLUGOwlME} & 9.4 & 13.2 & 10.2 & 7.3 &
35.1 & 10.1 & 12.5 & 10.2 & 8.9 &
38.2 & 10.5 & 21.3 & 10.5 & 10.3 & 39.4
\\

Video-LLaVA \cite{lin2023video} & 15.6 & 39.6 & 10.6 & 22.6 &
\underline{43.1} & 15.3 & 42.4 & 10.7 & 24.0 & 45.0 
& \underline{17.6} & 41.1 & 10.7 & 25.9 &
\underline{47.2} \\

mPLUG-Owl+V13B & \underline{15.7} & \underline{54.5} & 11.2 & \underline{26.9} &
42.5 & \underline{16.0} & \underline{62.3} & \underline{12.6} & \underline{28.6} &
\underline{46.0} & 17.0 & \underline{50.7} & \underline{12.8} & \underline{28.4} & 45.5
\\

 \midrule
VidTFS (V13B) & \textbf{23.0} & \textbf{80.1} & \textbf{15.4} & \textbf{32.3} & \textbf{47.6} 
& \textbf{23.1} & \textbf{91.7} & \textbf{16.9} & \textbf{35.0} &
\textbf{50.9} & \textbf{24.4} & \textbf{80.8} & \textbf{16.3} & \textbf{34.5} &
\textbf{50.2} \\
 

\hline\hline

 \multirow{2}{*}{\textbf{COIN}} &
 \multicolumn{5}{c|}{\begin{math}\rho\end{math} = 10\%} & 
 \multicolumn{5}{c|}{\begin{math}\rho\end{math} = 30\%} & 
 \multicolumn{5}{c}{\begin{math}\rho\end{math} = 50\%} \\
 
 & \multirow{1}{*}{\footnotesize{S}} & \multirow{1}{*}{\footnotesize{C}} 
 & \multirow{1}{*}{\footnotesize{M}} & \multirow{1}{*}{\footnotesize{B}} & \multirow{1}{*}{\footnotesize{SB}}   
 
 & \multirow{1}{*}{\footnotesize{S}} & \multirow{1}{*}{\footnotesize{C}} 
 & \multirow{1}{*}{\footnotesize{M}} & \multirow{1}{*}{\footnotesize{B}} & \multirow{1}{*}{\footnotesize{SB}} 
 
 & \multirow{1}{*}{\footnotesize{S}} & \multirow{1}{*}{\footnotesize{C}} 
 & \multirow{1}{*}{\footnotesize{M}} & \multirow{1}{*}{\footnotesize{B}} & \multirow{1}{*}{\footnotesize{SB}}\\

  \midrule
BLIP-2\cite{li2023blip} & 14.4 & 27.1 & 9.4 & 14.8 & 34.5
& 14.2 & 27.7 & 9.4 & 15.8 & 36.0
& 14.8 & 28.9 & 9.7 & 16.4 & 37.2
\\

InstructBLIP\cite{instructBLIP} & 7.0 & 11.6 & 6.4 & 3.7 & 27.6
& 6.8 & 9.4 & 6.0 & 4.0 & 27.7
& 7.6 & 10.6 & 6.5 & 4.2 & 28.3
\\

Video-ChatGPT\cite{videochatgpt} & 13.2 & 29.4 & 10.7 & 14.8 & 41.5
& 13.3 & 29.1 & 10.6 & 14.8 & 41.8
& 12.5 & 28.0 & 10.5 & 14.7 & 41.0
\\

mPLUG-Owl \cite{Ye2023mPLUGOwlME} & 10.8 & 15.4 & 8.7 & 7.6 & 35.7
& 11.8 & 18.9 & 9.7 & 9.4 & 40.0
& 12.8 & 21.4 & 10.5 & 10.3 & 42.2
\\

Video-LLaVA \cite{lin2023video} & \textbf{21.0} & 45.2 &  \underline{12.1} & 19.9 & \textbf{48.7}
& \underline{21.3} & 44.5 & \underline{12.0} & 20.2 & \underline{48.8}
& \underline{20.4} & 43.5 & 11.9 & 19.8 & \underline{48.1}
\\

mPLUG-Owl+V13B & 19.3 & \underline{60.3} & 11.9 & \textbf{28.6} & \underline{47.3} 
& 18.9 & \underline{61.2} & \underline{12.0} & \underline{29.0} & 47.5 
& 20.1 & \underline{63.7} & \underline{12.1} & \underline{29.3} & 47.7 
\\

 \midrule
VidTFS (V13B) & \underline{20.4} & \textbf{62.6} & \textbf{12.5} & \underline{27.2} & 45.0
& \textbf{23.0} & \textbf{71.4} & \textbf{13.7} & \textbf{30.4} & \textbf{49.6}
& \textbf{25.1} & \textbf{76.7} & \textbf{14.3} & \textbf{31.6} & \textbf{51.5}
\\

 \bottomrule
\end{tabular}
}
\caption{Open-vocabulary goal inferences results on CrossTask and COIN datasets. We report following metrics in \%: SPICE (S), CIDEr (C), METEOR (M), BERTScore (B), and SBERT (SB). Best and second best results are highlighted by bold and underline.}
\label{tab:open_ended_goal_inference}
\end{table}

For $\phi$ = goal inference task, we evaluate the VidTFS on CrossTask and COIN datasets. 
Specifically, we infer the person's goal with open vocabulary when chronologically observing the initial $\rho$=10\%, 30\%, and 50\% parts of the videos. 
As shown in Table \ref{tab:open_ended_goal_inference}, we observe that the VidTFS outperforms the current SOTA multimodal language models (MLMs) on most evaluation metrics under training-free open-vocabulary setting. Notably, the VidTFS surpassed pre-trained end-to-end MLMs, including the BLIP models, Video-LLaVA, and mPLUG-Owl. Compared with mPLUG-Owl + Vicuna which uses LLM to make inferences by using video-level description from MLM, VidTFS is outperformed it by +5.1 (47.6 \textit{vs} 42.5) at $\rho$=10\%, +4.9 (50.9 \textit{vs} 46.0) at $\rho$=30\% and +4.7 (50.2 \textit{vs} 45.5) at $\rho$=50\% respectively. This trend indicates that with the help of the frame selection module, VidTFS can make better inferences.
On the COIN dataset, with shorter inputs $\rho$=10\%, VidTFS fall behind Video-LLaVA by -3.7 (45.0 \textit{vs} 48.7) on SBERT; when $\rho$=30\%, VidTFS surpass the Video-LLaVA by +0.8 (49.6 \textit{vs} 48.8); and $\rho$=50\%, VidTFS broaden the gap by +3.4 (51.5 \textit{vs} 48.1). The proposed VidTFS shows improvement across the $\rho$ on the goal inference task in overall. 
The reason is that VidTFS can select relevant frames, thus effectively keeping the necessary information in long-duration untrimmed videos, whereas the other methods lack this flexibility.

\subsection{Action Recognition}
\label{sec:zsar}
\begin{table}[t]

\scriptsize
\centering
\begin{tabular}{l|ccccc|ccccc}
\toprule 
 \multirow{2}{*}{\textbf{Method}} & \multicolumn{5}{c|}{\textbf{UCF101}}& \multicolumn{5}{c}{\textbf{ActivityNet}}\\
 & \multirow{1}{*}{\footnotesize{S}} & \multirow{1}{*}{\footnotesize{C}} 
 & \multirow{1}{*}{\footnotesize{M}} & \multirow{1}{*}{\footnotesize{B}} & \multirow{1}{*}{\footnotesize{SB}}
  & \multirow{1}{*}{\footnotesize{S}} & \multirow{1}{*}{\footnotesize{C}} 
 & \multirow{1}{*}{\footnotesize{M}} & \multirow{1}{*}{\footnotesize{B}} & \multirow{1}{*}{\footnotesize{SB}}\\

 \midrule
BLIP-2\cite{li2023blip} & \textbf{21.0} & 48.9 & \textbf{16.2} & 12.5 & 60.6 & \underline{22.3} & 72.1 & 13.6 & 18.4 & 53.6\\

InstructBLIP\cite{instructBLIP} & \textbf{21.0} & \textbf{87.8} & 13.2 & 21.8 & \underline{61.9} & 10.3 & 42.2 & 6.5 & 5.5 & 46.5 \\

Video-ChatGPT\cite{videochatgpt} & 13.6 & 27.7 & 13.2 & 3.0 & 54.0 & 17.9 & 46.3 & 13.3 & 13.0 & 54.6 \\

mPLUG-Owl \cite{Ye2023mPLUGOwlME}& 13.4 & 31.7 & 13.9 & 5.8 & 54.8 & 14.8 & 33.0 & 11.5 & 11.0 & 51.1\\

Video-LLaVA \cite{lin2023video}& 12.1 & 24.8 & 12.7 & 4.9 & 50.2 & 19.8 & 47.6 & \textbf{14.9} & 16.6 & 53.7\\

mPLUG-Owl+V13B & 18.2 & 71.7 & 12.9 & \underline{24.5} & 58.7 & 22.0 & \underline{82.2} & 13.2 & \underline{25.6} & \underline{59.2}\\

 \midrule
VidTFS (V13B) & \underline{20.7} & \underline{83.9} & \underline{15.7}& \textbf{29.3} & \textbf{63.6} & \textbf{24.0} & \textbf{94.0} & \underline{14.7} & \textbf{28.8} & \textbf{61.0} \\

\bottomrule
\end{tabular}

\caption{Open-vocabulary action recognition on UCF101 and ActivityNet1.3 datasets.}
\label{tab:open_ended_action_recognition}
\end{table}

We validate the generalization of VidTFS on video action inference task (i.e., $\phi$ = action recognition). We test the VidTFS on UCF101 and ActivityNet datasets by using full untrimmed video length under new conditions of training-free and open-vocabulary settings. 
As in Table \ref{tab:open_ended_action_recognition}, we find that VidTFS outperforms SOTA multimodal LLM on UCF101 and ActivityNet datasets in terms of BERTScore and SBERT. 
This indicates that VidTFS could generate good semantically equivalent inference as the ground truth categories. 
However, in terms of metrics, such as SPICE, CIDEr and METEOR, the VidTFS falls behind models like BLIP-2 and InstructBLIP on UCF101 dataset. 
The reason is that the BLIPs are pre-trained on image-captioning tasks and excel at generating short image-level captions.
Besides, each frame from the short video of UCF101 is more likely to convey similar information about the actions and therefore frame selection may not be that important in those short videos. In contrast, VidTFS shows better performances on ActivityNet dataset which contained noisy video input which highlight the advantage of dynamic frame selection.
Since action videos contain fewer sub-events (steps) than long-duration instructional videos (e.g., CrossTask), VidTFS's advantage is lower than that of the goal inference task. 
However, we validate that compared with end-to-end pre-trained multimodal LLM, the VidTFS still achieves comparable performance. 
These findings validate the generalizability of VidTFS and its potential to be extended to other action-relevant tasks without training. 
\begin{table}[t]
\centering
\scriptsize
\resizebox{\textwidth}{!}{

\centering
\begin{tabular}{l|ccccc|ccccc|ccccc}
\toprule 
 \multirow{2}{*}{\textbf{CrossTask}} & 
 \multicolumn{5}{c|}{\begin{math}\rho\end{math} = 10\%} & 
 \multicolumn{5}{c|}{\begin{math}\rho\end{math} = 30\%} & 
 \multicolumn{5}{c}{\begin{math}\rho\end{math} = 50\%} \\
 
 & \multirow{1}{*}{\footnotesize{S}} & \multirow{1}{*}{\footnotesize{C}} 
 & \multirow{1}{*}{\footnotesize{M}} & \multirow{1}{*}{\footnotesize{B}} & \multirow{1}{*}{\footnotesize{SB}}   
 
 & \multirow{1}{*}{\footnotesize{S}} & \multirow{1}{*}{\footnotesize{C}} 
 & \multirow{1}{*}{\footnotesize{M}} & \multirow{1}{*}{\footnotesize{B}} & \multirow{1}{*}{\footnotesize{SB}} 
 
 & \multirow{1}{*}{\footnotesize{S}} & \multirow{1}{*}{\footnotesize{C}} 
 & \multirow{1}{*}{\footnotesize{M}} & \multirow{1}{*}{\footnotesize{B}} & \multirow{1}{*}{\footnotesize{SB}}\\
 \midrule
w/o ES & 18.3 & 61.3 & 12.7 & 25.0 & 42.9 
& 19.7 & 72.0 & 14.0 & 27.5 & 46.8 
& 22.1 & \textbf{83.0} & 15.1 & 30.3 & 48.8  
\\

\textbf{with ES}& \textbf{23.0} & \textbf{80.1} & \textbf{15.4} & \textbf{32.3} & \textbf{47.6} 
& \textbf{23.1} & \textbf{91.7} & \textbf{16.9} & \textbf{35.0} &
\textbf{50.9} & \textbf{24.4} & 80.8 & \textbf{16.3} & \textbf{34.5} &
\textbf{50.2}
\\

 \bottomrule
\end{tabular}

}

\vspace{5pt}
\resizebox{\textwidth}{!}{

\centering
\begin{tabular}{l|ccccc|ccccc|ccccc}
\toprule 
 \multirow{2}{*}{\textbf{COIN}} & 
 \multicolumn{5}{c|}{\begin{math}\rho\end{math} = 10\%} & 
 \multicolumn{5}{c|}{\begin{math}\rho\end{math} = 30\%} & 
 \multicolumn{5}{c}{\begin{math}\rho\end{math} = 50\%} \\
 
 & \multirow{1}{*}{\footnotesize{S}} & \multirow{1}{*}{\footnotesize{C}} 
 & \multirow{1}{*}{\footnotesize{M}} & \multirow{1}{*}{\footnotesize{B}} & \multirow{1}{*}{\footnotesize{SB}}   
 
 & \multirow{1}{*}{\footnotesize{S}} & \multirow{1}{*}{\footnotesize{C}} 
 & \multirow{1}{*}{\footnotesize{M}} & \multirow{1}{*}{\footnotesize{B}} & \multirow{1}{*}{\footnotesize{SB}} 
 
 & \multirow{1}{*}{\footnotesize{S}} & \multirow{1}{*}{\footnotesize{C}} 
 & \multirow{1}{*}{\footnotesize{M}} & \multirow{1}{*}{\footnotesize{B}} & \multirow{1}{*}{\footnotesize{SB}}\\
 \midrule
w/o ES & 18.3 & 52.8 & 11.4 & 23.0 & 41.9 
& 21.0 & 63.2 & 12.7 & 26.7 & 46.1
& 22.0 & 68.2 & 13.2 & 27.8 & 47.7
\\

\textbf{with ES}& \textbf{20.4} & \textbf{62.6} & \textbf{12.5} & \textbf{27.2} & \textbf{45.0}
& \textbf{23.0} & \textbf{71.4} & \textbf{13.7} & \textbf{30.4} & \textbf{49.6}
& \textbf{25.1} & \textbf{76.7} & \textbf{14.3} & \textbf{31.6} & \textbf{51.5}
\\

 \bottomrule
\end{tabular}

}


\vspace{5pt}

\centering
\begin{tabular}{l|ccccc}
\toprule 
 \multirow{2}{*}{\textbf{ActivityNet}} & 
 \multicolumn{5}{c}{\begin{math}\rho\end{math} = 100\%} \\

 & \multirow{1}{*}{\footnotesize{S}} & \multirow{1}{*}{\footnotesize{C}} 
 & \multirow{1}{*}{\footnotesize{M}} & \multirow{1}{*}{\footnotesize{B}} & \multirow{1}{*}{\footnotesize{SB}}\\
 \midrule
w/o ES & 21.2 & 79.6 & 12.7 & 22.3 & 57.4 \\

\textbf{with ES}& \textbf{24.0} & \textbf{94.0} & \textbf{14.7} & \textbf{28.8} & \textbf{61.0} \\
 \bottomrule

\end{tabular}


\caption{Ablation study of the evidence selector (ES) component across CrossTask, COIN and ActivityNet datasets.}
\label{tab:ablation_evidence_selector}
\end{table}

\subsection{Ablation Study}
\label{sec:ablation}
\paragraph{Ablation Evidence Selector Component.} We compare the performance of the VidTFS framework against a simple counterpart without an evidence selector.
This baseline uses BLIP-2 as a visual descriptor and  Vicuna13B to directly infer the goal or the action based on the given frame captions. 
The baseline does not generate steps (\cref{eq:step_1}) and there is no Evidence Selector Component.
As in Table \ref{tab:ablation_evidence_selector}, we observe that the performance drops without evidence selector for all three datasets on goal inference as well as action recognition.
We can conclude that the evidence selector helps to find relevant frames and helps to generate captions that support accurate goal inference and action recognition tasks.
Some qualitative results of selected frames are shown in the supplementary material demonstrating the correctness of the Evidence Selector.

\paragraph{Select Evidences from Visual Frames \textit{vs} Frame Captions}
We also investigate the effect of choosing relevant frames based on the original frame captions $\mathcal{C}$ and hypothesis steps using text-to-text matching.
We compare the steps $\mathcal{S}$ with frame-captions $\mathcal{C}$ using text-to-text similarity using SBERT model-based text embeddings. 
Then those frames (captions) with the highest similarity to the steps are selected.
We compare the step-to-caption approach vs the step-to-visual-frame similarity-based approach that uses CLIP visual features.
Results in
\Cref{tab:ablation_clip_sbert} show that the use of the CLIP model to select visual frames is better than using SBERT-based text matching. More ablations in supplementary.
\begin{table}[tb!]

\centering
\scriptsize{
\resizebox{\textwidth}{!}{
\begin{tabular}{l|ccccc|ccccc|ccccc}
\toprule 
 \multirow{2}{*}{\textbf{Method}} & 
 \multicolumn{5}{c|}{\begin{math}\rho\end{math} = 10\%} & 
 \multicolumn{5}{c|}{\begin{math}\rho\end{math} = 30\%} & 
 \multicolumn{5}{c}{\begin{math}\rho\end{math} = 50\%} \\
 
 & \multirow{1}{*}{\footnotesize{S}} & \multirow{1}{*}{\footnotesize{C}} 
 & \multirow{1}{*}{\footnotesize{M}} & \multirow{1}{*}{\footnotesize{B}} & \multirow{1}{*}{\footnotesize{SB}}   
 
 & \multirow{1}{*}{\footnotesize{S}} & \multirow{1}{*}{\footnotesize{C}} 
 & \multirow{1}{*}{\footnotesize{M}} & \multirow{1}{*}{\footnotesize{B}} & \multirow{1}{*}{\footnotesize{SB}} 
 
 & \multirow{1}{*}{\footnotesize{S}} & \multirow{1}{*}{\footnotesize{C}} 
 & \multirow{1}{*}{\footnotesize{M}} & \multirow{1}{*}{\footnotesize{B}} & \multirow{1}{*}{\footnotesize{SB}}\\

 \midrule

Steps-to-caption & 21.8 & 75.1 & 15.3 & \textbf{32.8} & 47.2
& 22.3 & \textbf{96.7} & 16.8 & \textbf{35.3} & 50.6
& 23.3 & \textbf{81.3} & 15.8 & 34.2 & 49.3
\\

Steps-to-frame(visual) & \textbf{23.0} & \textbf{80.1} & \textbf{15.4} & 32.3 & \textbf{47.6} 
& \textbf{23.1} & 91.7 & \textbf{16.9} & 35.0 &
\textbf{50.9} & \textbf{24.4} & 80.8 & \textbf{16.3} & \textbf{34.5} &
\textbf{50.2}
\\

 \bottomrule
\end{tabular}}
}
\centering
\caption{Comparison between step-to-frame vs step-to-caption matching in the Evidence Selector component on CrossTask dataset for goal inferences.}
\label{tab:ablation_clip_sbert}
\end{table}

\section{Discussion and Conclusion}

In conclusion, this work introduces the VidTFS, a training-free modular framework for open-vocabulary video goal inference and action recognition. The VidTFS use three frozen foundational models: BLIP-2, CLIP and Vicuna to accomplish four stage video inference process: \textit{See}, \textit{Guess}, \textit{Select}, and \textit{Infer}. 
We validate that these foundational models could play different roles and interact well with each other through language.
We also propose a training-free evidence selector that dynamically picks relevant frames for drawing inference. 
We experimentally verified that the VidTFS with dynamic frame selection module is effective and generalizable to two video inference tasks.
The VidTFS can be updated with more advanced foundational models to obtain even better results. 
VidTFS's limitations lie in its reliance on LLMs to draw inferences; thereby, it is difficult to control the generation process, and it suffers from LLM drawbacks like hallucinations. 
Besides, LLMs are statistical-based methods and do not contain an explicit logical reasoning process, causing VidTFS to have weak explainability.
Despite the current limitations, the proposed framework serves as a novel idea for training-free open-vocabulary inference tasks on video data.

\section{Acknowledgments}
This research / project is supported by the National Research Foundation, Singapore, under its NRF Fellowship (Award NRF-NRFF14-2022-0001). Any opinions, findings and conclusions or recommendations expressed in this material are those of the author(s) and do not reflect the views of National Research Foundation, Singapore.

\bibliographystyle{unsrtnat}
\bibliography{references}  

\newpage
\setcounter{page}{1}
\setcounter{section}{0}
\emergencystretch 3em

\section*{Supplementary Materials: Training-Free Action Recognition and Goal Inference with Dynamic Frame Selection}

The supplementary material is organized as follows:
Section (\ref{sec:additional_metrics}) discuss about additional evaluation other than conventional metrics;
Section (\ref{sec:add_impact_frame_selection}) presents additional analysis on the impact of frame selection;
Section (\ref{sec:add_analysis}) reports the supplemental ablations and analysis; Section (\ref{sec:implementation}) presents the implementation details of baselines and VidTFS including the prompts for large language models used in experiments;
Section (\ref{sec:time_inferences}) reports the inference time and number of LLM calls,
and lastly, Section (\ref{sec:qualitative}) presents some qualitative results that included more details of inference process examples.

\section{Additional Evaluation}
\label{sec:additional_metrics}
In addition to the conventional evaluation metrics such as METEOR(M)~\cite{denkowski2014meteor}, SPICE(S)~\cite{anderson2016spice}, CIDEr(C)~\cite{vedantam2015cider} that commonly used in visual captioning tasks \cite{anderson2018bottom}, as well as 
measuring semantic similarity by BERTScore(B)~\cite{zhang2019bertscore}, SBERT(SB)~\cite{Reimers2019SentenceBERTSE}), we use Llama3-8B\footnote {https://llama.meta.com/llama3/} model as a ``judge'' to compare the generated inferences with the ground truths. 
This is inspired by recent works that found the Large Language Model (LLM) can help act as a ``judge'' to determine the relevance of the inferences \cite{Bubeck2023SparksOA, Chan2023CLAIREI}.
Instead of letting the LLM provide a rating score to the generated inferences, we ask the LLM to provide binary output ``yes'' or ``no'' to determine whether both generated inference and ground truth have similar meanings.
The prompt for Llama3 is shown in Figure \ref{fig:prompt_llama3}.

\begin{figure}[h]
\begin{center}
\noindent\fbox{\begin{minipage}[h][2\height][c]{\dimexpr\textwidth-2\fboxsep-2\fboxrule\relax}
\texttt{Let A = <Ground Truth Label>, Let B = <Inferences>.\newline
Determine if A and B have similar meanings, then provide a binary output of 'Yes' or 'No' only.}
\end{minipage}}
\end{center}
   \caption{Prompt for Llama3 to judge correctness between the generated inferences and ground truth.}
\label{fig:prompt_llama3}
\end{figure}

\noindent
The results in Table \ref{tab:accuracy_by_llama3} show the generated inferences by VidTFS received more ``Yes'' ratio from Llama-3 judge than the other methods.

\begin{table}[ht]

\scriptsize
\centering
\begin{tabular}{l|ccc|ccc|c|c}
  \toprule
    \multirow{2}{*}{Method} & \multicolumn{3}{c|}{\textbf{CrossTask}}& \multicolumn{3}{c|}{\textbf{COIN}} & \textbf{UCF101} & \textbf{ActivityNet}\\

    & \multirow{1}{*}{10\%} 
    & \multirow{1}{*}{30\%} 
    & \multirow{1}{*}{50\%} 
    & \multirow{1}{*}{10\%} 
    & \multirow{1}{*}{30\%} 
    & \multirow{1}{*}{50\%}
    & \multirow{1}{*}{100\%}
    & \multirow{1}{*}{100\%}\\
    
    \midrule
    BLIP-2\cite{li2023blip}& 32.2 & 34.1 & 35.8 & 31.2 & 31.6 & 32.2 & 72.8 & 53.4\\
    InstructBLIP\cite{instructBLIP}& 11.7 & 10.0 & 10.4 & 16.1 & 15.1 & 14.8 & \underline{74.8} & 54.1\\
    Video-ChatGPT\cite{videochatgpt}& 22.4 & 19.8 & 21.0 & 24.6 & 25.0 & 24.3 & 64.7 & 44.7\\
    mPLUG-Owl \cite{Ye2023mPLUGOwlME}& 27.8 & 38.8 & 42.8 & 26.8 & 32.1 & 34.6 & 65.9 & 49.0\\
    Video-LLaVA \cite{lin2023video}& \underline{42.2} & \underline{43.6} & \underline{49.0} & \textbf{42.5} & \textbf{43.0} & \underline{41.2} & 63.6 & \underline{60.4}\\
    mPLUG-Owl+V13B & 39.1 & 43.1 & 44.5 & \underline{38.7} & 38.6 & 30.4 & 74.1 & 54.2\\
    \midrule
    VidTFS (V13B) & \textbf{51.8} & \textbf{58.1} & \textbf{63.2} & 38.1 & \underline{42.5} & \textbf{47.3} & \textbf{79.7} & \textbf{71.9}\\
    \bottomrule
  \end{tabular}

\caption{Accuracy evaluated by Llama3 judge. Best and second best results are highlighted by bold and underline respectively.}
\label{tab:accuracy_by_llama3}
\end{table}

\section{Additional Analysis on Impact of Frame Selection}
\label{sec:add_impact_frame_selection}
To further evaluate the effectiveness of our evidence selector, we measured how well the selected frames matched the ground truth label. We use different frame sampling methods for frame selection; we then use CLIP\cite{clip} to calculate the similarity between the selected visual frame and text labels.
We obtained the visual features by averaging the sampled frames. The results shown in Table \ref{tab:sim_score_before_after} indicate that the frames selected by VidTFS evidence selector have better similarity scores with the text features of the ground truth label.
\begin{table}[h]

\scriptsize
\centering
\begin{tabular}{l|ccc|ccc|c}
  \toprule
    \multirow{2}{*}{Method} & \multicolumn{3}{c|}{\textbf{CrossTask}}& \multicolumn{3}{c|}{\textbf{COIN}} & \textbf{ActivityNet}\\

    & \multirow{1}{*}{10\%} 
    & \multirow{1}{*}{30\%} 
    & \multirow{1}{*}{50\%} 
    & \multirow{1}{*}{10\%} 
    & \multirow{1}{*}{30\%} 
    & \multirow{1}{*}{50\%}
    & \multirow{1}{*}{100\%}\\
    
    \midrule
    Uniformly sampled & 0.764 & 0.780 & 0.788 & 0.768 & 0.793 & 0.800 & 0.815\\
    Randomly sampled & 0.759 & 0.777 & 0.783 & 0.763 & 0.789 & 0.796 & 0.813\\
    \midrule
    VidTFS dynamic sampled & \textbf{0.784} & \textbf{0.802} & \textbf{0.806} & \textbf{0.781} & \textbf{0.802} & \textbf{0.818} & \textbf{0.831}\\
    \bottomrule
  \end{tabular}

\caption{Similarity score between visual and text features by CLIP after frame selection process.}
\label{tab:sim_score_before_after}
\end{table}

\section{Supplemental Ablations and Results}
\label{sec:add_analysis}
\noindent
\subsection{Select Evidence using Hypotheses versus Expanded Hypothesized Steps by LLM}
We compare with counterparts that directly use top-$k$ hypotheses, $\mathcal{H}$, to select the relevant frames from the $N$ sampled frames. Table \ref{tab:ablation_use_hypo_find_frame} shows that using LLM-generated steps to find the evidence frames is better for inference performance. 

\begin{table}[h]

\centering
\scriptsize{
\resizebox{\textwidth}{!}{
\begin{tabular}{l|ccccc|ccccc|ccccc}
\toprule 
 \multirow{2}{*}{\textbf{Method}} & 
 \multicolumn{5}{c|}{\begin{math}\rho\end{math} = 10\%} & 
 \multicolumn{5}{c|}{\begin{math}\rho\end{math} = 30\%} & 
 \multicolumn{5}{c}{\begin{math}\rho\end{math} = 50\%} \\
 
 & \multirow{1}{*}{\footnotesize{S}} & \multirow{1}{*}{\footnotesize{C}} 
 & \multirow{1}{*}{\footnotesize{M}} & \multirow{1}{*}{\footnotesize{B}} & \multirow{1}{*}{\footnotesize{SB}}   
 
 & \multirow{1}{*}{\footnotesize{S}} & \multirow{1}{*}{\footnotesize{C}} 
 & \multirow{1}{*}{\footnotesize{M}} & \multirow{1}{*}{\footnotesize{B}} & \multirow{1}{*}{\footnotesize{SB}} 
 
 & \multirow{1}{*}{\footnotesize{S}} & \multirow{1}{*}{\footnotesize{C}} 
 & \multirow{1}{*}{\footnotesize{M}} & \multirow{1}{*}{\footnotesize{B}} & \multirow{1}{*}{\footnotesize{SB}}\\

 \midrule

Use hypotheses & 21.9 & 79.9 & 15.2 & 31.9 & 46.9
& 21.6 & 84.2 & 16.4 & 33.8 & 49.5
& 23.6 & 79.9 & 16.2 & 33.6 & 50.1
\\

Use generated steps & \textbf{23.0} & \textbf{80.1} & \textbf{15.4} & \textbf{32.3} & \textbf{47.6} 
& \textbf{23.1} & \textbf{91.7} & \textbf{16.9} & \textbf{35.0} &
\textbf{50.9} & \textbf{24.4} & \textbf{80.8} & \textbf{16.3} & \textbf{34.5} &
\textbf{50.2}
\\

 \bottomrule
\end{tabular}}
}
\centering
\caption{Comparison between hypotheses-to-frame versus steps-to-frame matching in the Evidence Selector component on CrossTask dataset for goal inferences.}
\label{tab:ablation_use_hypo_find_frame}
\end{table}

\subsection{Select Evidence using Frame Captions versus Hypothesized Steps by LLM}
We also compare with counterparts that use frame captions $\mathcal{C}$ generated by visual descriptor (e.g. BLIP-2), and then use CLIP to select the relevant frames from the $N$ sampled frames. Table \ref{tab:ablation_use_captions_select_frames} shows that using LLM-generated steps to find the evidence frames is better for inference performance. 
\begin{table}[h]

\centering
\resizebox{\textwidth}{!}{\begin{tabular}{l|ccccc|ccccc|ccccc}
\toprule 
 \multirow{2}{*}{\textbf{Method}} & 
 \multicolumn{5}{c|}{\begin{math}\rho\end{math} = 10\%} & 
 \multicolumn{5}{c|}{\begin{math}\rho\end{math} = 30\%} & 
 \multicolumn{5}{c}{\begin{math}\rho\end{math} = 50\%} \\
 
 & \multirow{1}{*}{\footnotesize{S}} & \multirow{1}{*}{\footnotesize{C}} 
 & \multirow{1}{*}{\footnotesize{M}} & \multirow{1}{*}{\footnotesize{B}} & \multirow{1}{*}{\footnotesize{SB}}   
 
 & \multirow{1}{*}{\footnotesize{S}} & \multirow{1}{*}{\footnotesize{C}} 
 & \multirow{1}{*}{\footnotesize{M}} & \multirow{1}{*}{\footnotesize{B}} & \multirow{1}{*}{\footnotesize{SB}} 
 
 & \multirow{1}{*}{\footnotesize{S}} & \multirow{1}{*}{\footnotesize{C}} 
 & \multirow{1}{*}{\footnotesize{M}} & \multirow{1}{*}{\footnotesize{B}} & \multirow{1}{*}{\footnotesize{SB}}\\

 \midrule

Use captions & 21.4 & 79.7 & 15.1 & 31.2 & 45.9
& 21.4 & 83.9 & 16.7 & 33.3 & 48.9
& 22.2 & 80.4 & 15.8 & 33.3 & 49.3
\\

Use generated steps & \textbf{23.0} & \textbf{80.1} & \textbf{15.4} & \textbf{32.3} & \textbf{47.6} 
& \textbf{23.1} & \textbf{91.7} & \textbf{16.9} & \textbf{35.0} &
\textbf{50.9} & \textbf{24.4} & \textbf{80.8} & \textbf{16.3} & \textbf{34.5} &
\textbf{50.2}
\\

 \bottomrule
\end{tabular}
}

\centering
\caption{Comparison between captions-to-frame versus steps-to-frame matching in the Evidence Selector on CrossTask dataset for goal inference.}
\label{tab:ablation_use_captions_select_frames}
\end{table}

\noindent
\subsection{Ablation Number of Iteration of Frame Selection} 
We compare VidTFS (1 iteration) with a counterpart that perform 2 and 3 iterations of frame selection process. The Table \ref{tab:ablation_more_iteration} shows that more iterations of frame selection does not yield improvements. 
This reflects that one evidence selector is sufficient to select relevant frames for make inference and balance computations and performance well.
\begin{table}[h]

\centering
\scriptsize
\resizebox{\textwidth}{!}{
\begin{tabular}{l|ccccc|ccccc|ccccc}
\toprule 
 \multirow{2}{*}{\textbf{Method}} & 
 \multicolumn{5}{c|}{\begin{math}\rho\end{math} = 10\%} & 
 \multicolumn{5}{c|}{\begin{math}\rho\end{math} = 30\%} & 
 \multicolumn{5}{c}{\begin{math}\rho\end{math} = 50\%} \\
 
 & \multirow{1}{*}{\footnotesize{S}} & \multirow{1}{*}{\footnotesize{C}} 
 & \multirow{1}{*}{\footnotesize{M}} & \multirow{1}{*}{\footnotesize{B}} & \multirow{1}{*}{\footnotesize{SB}}   
 
 & \multirow{1}{*}{\footnotesize{S}} & \multirow{1}{*}{\footnotesize{C}} 
 & \multirow{1}{*}{\footnotesize{M}} & \multirow{1}{*}{\footnotesize{B}} & \multirow{1}{*}{\footnotesize{SB}} 
 
 & \multirow{1}{*}{\footnotesize{S}} & \multirow{1}{*}{\footnotesize{C}} 
 & \multirow{1}{*}{\footnotesize{M}} & \multirow{1}{*}{\footnotesize{B}} & \multirow{1}{*}{\footnotesize{SB}}\\
 \midrule

1 iteration & 23.0 & 80.1 & 15.4 & 32.3 & 47.6 
& 23.1 & 91.7 & 16.9 & 35.0 &
50.9 & 24.4 & 80.8 & 16.3 & 34.5 &
50.2 
\\

2 iterations & 23.1 & 73.6 & 15.0 & 33.3 & 47.5
& 21.8 & 76.2 & 15.8 & 33.4 & 49.2
& 23.4 & 83.2 & 16.1 & 32.5 & 49.4
\\

3 iterations & 23.5 & 74.6 & 15.4 & 32.8 & 47.6
& 20.7 & 72.4 & 15.2 & 32.7 & 48.6
& 22.9 & 80.3 & 16.2 & 33.5 &  49.7
\\

 \bottomrule
\end{tabular}
}
\centering
\caption{Ablation study on iteration of frame selection.} 
\label{tab:ablation_more_iteration}
\end{table}

\noindent
\subsection{Ablation Number of Frames.}
We also study the influence of the number of sampled frames, $L$, and selected frames, $M$ together, by varying the frame number limit so that $L, M\le\{4,8,16,32\}$. Table \ref{tab:ablation_num_frames_crosstask} shows that performance is optimal when limited to 16 frames, as it also indicates that including more frames does not improve performance.
\begin{table}[h]
\centering
\scriptsize{
\resizebox{\textwidth}{!}{\begin{tabular}{l|ccccc|ccccc|ccccc}
\toprule 
 \multirow{2}{*}{\textbf{Method}} & 
 \multicolumn{5}{c|}{\begin{math}\rho\end{math} = 10\%} & 
 \multicolumn{5}{c|}{\begin{math}\rho\end{math} = 30\%} & 
 \multicolumn{5}{c}{\begin{math}\rho\end{math} = 50\%} \\
 
 & \multirow{1}{*}{\footnotesize{S}} & \multirow{1}{*}{\footnotesize{C}} 
 & \multirow{1}{*}{\footnotesize{M}} & \multirow{1}{*}{\footnotesize{B}} & \multirow{1}{*}{\footnotesize{SB}}   
 
 & \multirow{1}{*}{\footnotesize{S}} & \multirow{1}{*}{\footnotesize{C}} 
 & \multirow{1}{*}{\footnotesize{M}} & \multirow{1}{*}{\footnotesize{B}} & \multirow{1}{*}{\footnotesize{SB}} 
 
 & \multirow{1}{*}{\footnotesize{S}} & \multirow{1}{*}{\footnotesize{C}} 
 & \multirow{1}{*}{\footnotesize{M}} & \multirow{1}{*}{\footnotesize{B}} & \multirow{1}{*}{\footnotesize{SB}}\\
 \midrule
4 frames & 19.1 & 59.5 & 12.9 & 29.4 & 43.3
& 16.8 & 68.6 & 13.2 & 30.2 & 44.0
& 16.5 & 69.6 & 13.1 & 31.6 & 45.5
\\

8 frames & \underline{20.4} & \underline{70.8} & 13.7 & 30.7 & 46.2
& \underline{21.1} & \underline{82.8} & \underline{15.6} & \underline{33.6} & \underline{49.6}
& 22.7 & \textbf{84.7} & 16.2 & \textbf{35.7} & \underline{50.8}
\\

16 frames & \textbf{23.0} & \textbf{80.1} & \textbf{15.4} & \textbf{32.3} & \textbf{47.6} 
& \textbf{23.1} & \textbf{91.7} & \textbf{16.9} & \textbf{35.0} &
\textbf{50.9} & \textbf{24.4} & 80.8 & \underline{16.3} & \underline{34.5} &
50.2 
\\

32 frames & 19.3 & 64.0 & \underline{14.8} & \underline{31.1} & \underline{46.4}
& 21.0 & 79.9 & 15.5 & 30.7 & 47.3 
& \underline{23.5} & \underline{83.8} & \textbf{17.1} & \underline{34.5} & \textbf{51.5}
\\
 \bottomrule
\end{tabular}}

}
\caption{Ablation of number of sampled frames ($L$) and relevant frames selected ($M$).}
\label{tab:ablation_num_frames_crosstask}
\end{table}

\noindent
\subsection{Ablation on Large Language Model.}
We conduct ablation on using different LLM (e.g. Vicuna\cite{zheng2023judging}, GPT-3.5-Turbo \cite{gpt3}, Llama-3-8B-Instruct) in the $\mathcal{F}_{\text{LLM}}$ and compare their inference performance. 
As shown in Table \ref{tab:ablation_llm}, the Vicuna-13B model performs better than Vicuna-7B while achieving comparable performance with GPT-3.5.
In addition, we also compared with the quantized Vicuna-13B-8bit model and Vicuna-13B model from \cite{2023lmdeploy} which compresses the LLM and speeds up the inference as shown in Section \ref{sec:time_inferences}. 
This ablation study suggests that using more robust LLMs could enhance inference performance.
\begin{table*}[h]
\centering
\scriptsize{
\resizebox{\textwidth}{!}{\begin{tabular}{l|ccccc|ccccc|ccccc}
\toprule 
 \multirow{2}{*}{\textbf{Method}} & 
 \multicolumn{5}{c|}{\begin{math}\rho\end{math} = 10\%} & 
 \multicolumn{5}{c|}{\begin{math}\rho\end{math} = 30\%} & 
 \multicolumn{5}{c}{\begin{math}\rho\end{math} = 50\%} \\
 
 & \multirow{1}{*}{\footnotesize{S}} & \multirow{1}{*}{\footnotesize{C}} 
 & \multirow{1}{*}{\footnotesize{M}} & \multirow{1}{*}{\footnotesize{B}} & \multirow{1}{*}{\footnotesize{SB}}   
 
 & \multirow{1}{*}{\footnotesize{S}} & \multirow{1}{*}{\footnotesize{C}} 
 & \multirow{1}{*}{\footnotesize{M}} & \multirow{1}{*}{\footnotesize{B}} & \multirow{1}{*}{\footnotesize{SB}} 
 
 & \multirow{1}{*}{\footnotesize{S}} & \multirow{1}{*}{\footnotesize{C}} 
 & \multirow{1}{*}{\footnotesize{M}} & \multirow{1}{*}{\footnotesize{B}} & \multirow{1}{*}{\footnotesize{SB}}\\

 \midrule

Vicuna (7B) & 20.1 & 77.2 & 13.4 & 30.5 & 45.4
& 21.5 & 88.6 & 14.3 & 32.0 & 47.4
& 21.2 & 86.5 & 14.8 & 32.6 & 48.6
\\

Vicuna (13B) & \underline{23.0} & \textbf{80.1} & 15.4 & 32.3 & 47.6
& \textbf{23.1} & 91.7 & 16.9 & 35.0 &
\underline{50.9} & \textbf{24.4} & 80.8 & 16.3 & 34.5 &
50.2 
\\

Vicuna (13B) by \cite{2023lmdeploy} & \textbf{23.8} & \underline{78.6} & \underline{15.6} & \underline{33.5} & \underline{48.3} 
& 21.3 & 82.9 & 15.7 & 33.3 & 49.4 
& 22.7 & 76.1 & 16.0 & 33.0 & 49.6
\\

Vicuna (13B) 8bit & 21.0 & 74.9 & \textbf{16.8} & \textbf{34.2} & \textbf{48.9} 
& 20.7 & 80.6 & \underline{17.1} & 35.2 & 50.7 
& \underline{23.9} & 82.5 & 17.0 & 36.5 & 51.5
\\

GPT-3.5-Turbo & 18.7 & 75.4 & 15.5 & 31.3 & 47.0
& 19.6 & \underline{92.3} & 16.7 & \underline{35.5} & \textbf{51.3} 
& 20.9 & \underline{88.6} & \underline{17.5} & \underline{37.8} & \textbf{52.5}
\\

Llama3 (8B) & 18.8 & 75.4 & 15.4 & 29.8 & 44.6 
& \underline{21.9} & \textbf{109.3} & \textbf{18.0} & \textbf{37.6} & \textbf{51.3}
& 23.3 & \textbf{116.9} & \textbf{17.9} & \textbf{40.4} & \underline{51.7}
\\

 \bottomrule
\end{tabular}}
}
\caption{Ablation study of the LLMs.} 
\label{tab:ablation_llm}
\end{table*}

\noindent
\subsection{In-Context Learning Prompt.}
We ablate the effect of In-Context Learning~\cite{gpt3,min2022rethinking,rubin2021learning} (ICL) within the LLM prompt for open-vocabulary inference in the LLM prompt. Table \ref{tab:ablation_icl} results suggest that using ICL helps improve open-vocabulary inference performance. 
\begin{table}[h]

\centering
\scriptsize{
\resizebox{\textwidth}{!}{\begin{tabular}{l|ccccc|ccccc|ccccc}
\toprule 
 \multirow{2}{*}{\textbf{Method}} & 
 \multicolumn{5}{c|}{\begin{math}\rho\end{math} = 10\%} & 
 \multicolumn{5}{c|}{\begin{math}\rho\end{math} = 30\%} & 
 \multicolumn{5}{c}{\begin{math}\rho\end{math} = 50\%} \\
 
 & \multirow{1}{*}{\footnotesize{S}} & \multirow{1}{*}{\footnotesize{C}} 
 & \multirow{1}{*}{\footnotesize{M}} & \multirow{1}{*}{\footnotesize{B}} & \multirow{1}{*}{\footnotesize{SB}}   
 
 & \multirow{1}{*}{\footnotesize{S}} & \multirow{1}{*}{\footnotesize{C}} 
 & \multirow{1}{*}{\footnotesize{M}} & \multirow{1}{*}{\footnotesize{B}} & \multirow{1}{*}{\footnotesize{SB}} 
 
 & \multirow{1}{*}{\footnotesize{S}} & \multirow{1}{*}{\footnotesize{C}} 
 & \multirow{1}{*}{\footnotesize{M}} & \multirow{1}{*}{\footnotesize{B}} & \multirow{1}{*}{\footnotesize{SB}}\\

 \midrule

without ICL & 19.7 & 46.4 & 12.1 & 19.0 & 42.4
& 18.9 & 38.2 & 11.9 & 16.7 & 42.3
& 18.5 & 36.3 & 11.2 & 16.1 & 41.8
\\

\textbf{with ICL} & \textbf{23.0} & \textbf{80.1} & \textbf{15.4} & \textbf{32.3} & \textbf{47.6} 
& \textbf{23.1} & \textbf{91.7} & \textbf{16.9} & \textbf{35.0} &
\textbf{50.9} & \textbf{24.4} & \textbf{80.8} & \textbf{16.3} & \textbf{34.5} &
\textbf{50.2} 
\\

 \bottomrule
\end{tabular}}
}

\centering
\caption{Ablation study of the In-Context Learning (ICL) prompt.}
\label{tab:ablation_icl}
\end{table}

\subsection{Hypothesis from CLIP.}
We also study the impact of the hypothesis $h_c$ from CLIP for video inference. The Table \ref{tab:ablation_hc} shows using ($\mathcal{H}$ $\oplus$ $\ddot{\mathcal{H}}$ $\oplus$ $h_c$) as an option list for the final stage inference brings a slight improvements.
\begin{table}[h]

\centering
\scriptsize{
\resizebox{\textwidth}{!}{\begin{tabular}{l|ccccc|ccccc|ccccc}
\toprule 
 \multirow{2}{*}{\textbf{Method}} & 
 \multicolumn{5}{c|}{\begin{math}\rho\end{math} = 10\%} & 
 \multicolumn{5}{c|}{\begin{math}\rho\end{math} = 30\%} & 
 \multicolumn{5}{c}{\begin{math}\rho\end{math} = 50\%} \\
 
 & \multirow{1}{*}{\footnotesize{S}} & \multirow{1}{*}{\footnotesize{C}} 
 & \multirow{1}{*}{\footnotesize{M}} & \multirow{1}{*}{\footnotesize{B}} & \multirow{1}{*}{\footnotesize{SB}}   
 
 & \multirow{1}{*}{\footnotesize{S}} & \multirow{1}{*}{\footnotesize{C}} 
 & \multirow{1}{*}{\footnotesize{M}} & \multirow{1}{*}{\footnotesize{B}} & \multirow{1}{*}{\footnotesize{SB}} 
 
 & \multirow{1}{*}{\footnotesize{S}} & \multirow{1}{*}{\footnotesize{C}} 
 & \multirow{1}{*}{\footnotesize{M}} & \multirow{1}{*}{\footnotesize{B}} & \multirow{1}{*}{\footnotesize{SB}}\\

 \midrule

w/o $h_c$ & 22.7 & 80.1 & 15.2 & 32.3 & 47.2
& 22.4 & 91.7 & 16.5 & 34.5 & 50.3
& 23.7 & 76.2 & 15.9 & 33.8 & 49.2
\\

With $h_c$ & \textbf{23.0} & 80.1 & \textbf{15.4} & 32.3 & \textbf{47.6} 
& \textbf{23.1} & 91.7 & \textbf{16.9} & \textbf{35.0} &
\textbf{50.9} & \textbf{24.4} & \textbf{80.8} & \textbf{16.3} & \textbf{34.5} &
\textbf{50.2} \\

 \bottomrule
\end{tabular}}
}

\centering
\caption{Ablation study of hypothesis from CLIP ($h_c$).}
\label{tab:ablation_hc}
\end{table}


\subsection{Operators to Combine Hypotheses List.}
\label{sec:ablation_operators}
We test two types of operators $\oplus$ to combine $\mathcal{H}$, $\ddot{\mathcal{H}}$ and $h_c$. 
One is list concatenation: 
[ $\mathcal{H}$ ] + [ $\ddot{\mathcal{H}}$ ] + [ $h_c$] 
and another is union of set \{ $\mathcal{H}$ \} $\cup$ \{ $\ddot{\mathcal{H}}$ \} $\cup$ \{$h_c$\}. 
Their main difference is list concatenation allows redundant options, but the union operator does not; this would affect the frequency of individual hypotheses inputted to LLM. As in Table \ref{tab:ablation_concatenation}, the concatenation operator performs better than the union operator.
\begin{table}[ht]

\centering
\scriptsize{
\resizebox{\textwidth}{!}{\begin{tabular}{l|ccccc|ccccc|ccccc}
\toprule 
 \multirow{2}{*}{\textbf{Method}} & 
 \multicolumn{5}{c|}{\begin{math}\rho\end{math} = 10\%} & 
 \multicolumn{5}{c|}{\begin{math}\rho\end{math} = 30\%} & 
 \multicolumn{5}{c}{\begin{math}\rho\end{math} = 50\%} \\
 
 & \multirow{1}{*}{\footnotesize{S}} & \multirow{1}{*}{\footnotesize{C}} 
 & \multirow{1}{*}{\footnotesize{M}} & \multirow{1}{*}{\footnotesize{B}} & \multirow{1}{*}{\footnotesize{SB}}   
 
 & \multirow{1}{*}{\footnotesize{S}} & \multirow{1}{*}{\footnotesize{C}} 
 & \multirow{1}{*}{\footnotesize{M}} & \multirow{1}{*}{\footnotesize{B}} & \multirow{1}{*}{\footnotesize{SB}} 
 
 & \multirow{1}{*}{\footnotesize{S}} & \multirow{1}{*}{\footnotesize{C}} 
 & \multirow{1}{*}{\footnotesize{M}} & \multirow{1}{*}{\footnotesize{B}} & \multirow{1}{*}{\footnotesize{SB}}\\

 \midrule

Set Union Operator & 22.8 & 77.1 & 15.4 & 31.8 & 47.2
& 21.8 & 83.0 & 15.8 & 33.2 & 49.5
& 23.4 & 78.2 & 15.9 & 33.8 & 49.8
\\

List concatenation & \textbf{23.0} & \textbf{80.1} & 15.4 & \textbf{32.3} & \textbf{47.6} 
& \textbf{23.1} & \textbf{91.7} & \textbf{16.9} & \textbf{35.0} &
\textbf{50.9} & \textbf{24.4} & \textbf{80.8} & \textbf{16.3} & \textbf{34.5} &
\textbf{50.2}
\\

 \bottomrule
\end{tabular}}
}

\centering
\caption{Ablation study on concatenation of hypotheses.}
\label{tab:ablation_concatenation}
\end{table}

\section{Implementation Details}
\label{sec:implementation}
In this section, we provide the implementation details of both baselines and the proposed VidTFS framework, including the prompts used to query the multimodal language model (MLM) and large language model (LLM).

\subsection{Open-vocabulary Inference Baselines}
\subsubsection{BLIP-2}
\label{sec:blip2_implementation}
BLIP-2 \cite{li2023blip} has proficient zero-shot image question-answering ability; we use it for frame-level inference (16 frames) as it is designed for image-to-text tasks. We use BLIP-2 with FLanT5-XXL model with the prompts: \texttt{``Question: What is the intention or goal of the person in the photo? Short answer: ''} for goal inference task, while \texttt{``Question: What is the ongoing action of the person in the photo? Short answer: ''} for the action recognition task. We then computed the evaluation metrics of each frame-level caption against the ground truth label and took the mean values as the final measurement of each video-level inference.

  

\subsubsection{InstructBLIP}
\label{sec:instructBLIP_implementation}
InstructBLIP \cite{instructBLIP} with FLanT5-XXL model is instruction-tuned based on pre-trained BLIP-2 \cite{li2023blip}. Instead of a question-answer format, we use an instruction format prompts:  \texttt{``Please provide the intention or goal of the person in the photo.''} for goal inference task, whereas \texttt{``Please provide a short answer of the ongoing action of the person in the photo.''} for the action recognition task. We use the same evaluation method as the BLIP-2 baseline since both are applied for frame-level inference (16 frames).

\subsubsection{Video-ChatGPT}
\label{sec:videochatgpt_implementation}
Video-ChatGPT \cite{videochatgpt} is pre-trained on 100K video-caption pairs and works well in various open-vocabulary video question-answering tasks. We provide the direct and clear question prompt, \texttt{``What is the intention or goal of the person in the video?''} and \texttt{``What is the ongoing action of the person in the video?''} to the model for zero-shot video goal inference and action recognition, respectively. We set the frame number parameter to 16.

\subsubsection{mPLUG-Owl}
\label{sec:mplugowl_implementation}
mPLUG-Owl \cite{Ye2023mPLUGOwlME} is another large MLM demonstrating remarkable zero-shot abilities on various open-vocabulary visual inference tasks. We follow the suggested prompt template, \texttt{`{}``The following is a conversation between a curious human and an AI assistant. The assistant gives helpful, detailed, and polite answers to the user's questions.
Human: <|video|>
Human: \{Question\}
AI: '{}''}. The \texttt{Question} is filled with \texttt{``What is the intention or goal of the person in the video?''} for the goal inference task, whereas \texttt{``What is the ongoing action of the person in the video?''} for the action recognition task. The number of sampled frames per video is 16.

\subsubsection{Video-LLaVA}
Video-LLaVA \cite{lin2023video} proposed as MLM that uses a unified visual representation before projection to enhance downstream visual-language understanding. We use it as a baseline to perform open-vocabulary video inference with the following prompts: \texttt{``Write a short answer of the intention or goal of the person in the video. The person in the video is: ''} for goal inference, whereas \texttt{``Write a short answer of the ongoing action of the person in the video. The person in the video is: ''} for action recognition. It is only supporting to take a maximum of 8 frames for each video inference at the moment we implemented it.

\subsubsection{Combination of mPLUG-Owl \& Vicuna-13B}
mPLUG-Owl + Vicuna-13B is another baseline method that use the mPLUG-Owl as a visual descriptor and Vicuna-13B as LLM agent to make inference without any frame selection process. We input the prompt to mPLUG-Owl as \texttt{`{}``The following is a conversation between a curious human and AI assistant. The assistant gives helpful, detailed, and polite answers to the user's questions.
Human: <|video|>
Human: What is the content of the video?
AI: '{}''}, and then we use the LLM to infer directly on top of the video description generated by mPLUG-Owl. The prompt for LLM is similar to the prompt template used by VidTFS as shown in Table \ref{tab:llm_prompt8_template}. Instead of list the top-k hypotheses, we ask the LLM to provide only one answer.

\subsection{VidTFS Framework}
\label{sec:vidtfs_detail}

\subsubsection{Seeing through Visual Descriptor.}
\label{sec:vd_implementation}
We use BLIP-2 with FLanT5-XXL \cite{li2023blip} to generate a caption for every sampled frame by using a general prompt ($\varphi_d$): \texttt{``Question: What is the content of the image? Answer: ''} for all inference tasks. After $L$ number of captions are generated, we preprocess the captions by deduplicate the identical captions if there is any and concatenate the rest by using the word ``\textit{then}'' to create a high-level description so that $\mathcal{D}$ follows the form of ``\texttt{<caption 1>, then, <catpion 2>, then, ... <caption L>}''. In a later process, we also do the same for the $M$ selected frames to generate a new description $\ddot{\mathcal{D}}$.

\subsubsection{Dynamic Frame Selection by Evidence Selector.}
\label{sec:es_implementation}
The evidence selector module is pivotal in aligning visual features with text features to identify the most relevant frames. We employ the frozen visual and text towers from the CLIP \cite{clip} model by using the ViT-B/16 backbone to effectively integrate visual and textual information for optimal evidence frame selection.
Specifically, we use CLIP vision encoder to encode $N$ visual frames and generate the frame features, then we use CLIP text encoder to generate text features by encoding the hypothesized steps $S$ generated by the LLM. 
Subsequently, we compute similarity between visual features and text features.
We select the top similarity score of $M$ frames and resulting in a new set of evidence frames.

\subsubsection{Guessing Hypotheses and Final Inference by LLM.}
\label{sec:llm_implementation}
We use the readily available LLMs, specifically Vicuna-13B \cite{vicuna2023}, in the goal inference and action recognition experiments. 
For Vicuna, we set the temperature to 0.001 and the repetition penalty to 1.0.
The full prompt template ($\varphi_v, \varphi_l, \varphi_f$) that are used to generate hypotheses ($\mathcal{H}$ or $\mathcal{\ddot{H}}$), hypothesized step sequence ($\mathcal{S}$), and final inference ($h$) are shown in Table \ref{tab:llm_prompt8_template}.
The prompt template is applied to both goal inference and action recognition tasks without requiring crafting the prompt again from task to task.

\section{Inferences Time and Number of LLM Calls}
\label{sec:time_inferences}
We record the inference time and number of LLM calls for comparison.
We tested all methods on a single NVIDIA A100 GPU using 10 videos.
The average time taken excludes the time required for loading and pre-processing the videos or visual frame, only start timing when prompting the model to make an inference based on a given inference task $\phi$ (e.g., ``goal inference'').
For BLIP-2 and InstructBLIP, we query the language model 16 times as we use them for frame-level inferences.
For mPLUG-Owl + Vicuna-13B, we only time the inference after mPLUG-Owl generate the video-level caption.
The proposed VidTFS that using original Vicuna-13B \cite{vicuna2023} shows a longer inference time compared to the multimodal language models (MLMs) which only need one LLM call.
However, the inference time of VidTFS could potentially be shortened through engineering efforts, as shown by using the quantized model, or LLM from \cite{2023lmdeploy}, which compresses and serves LLM more efficiently, but resulting in degraded inference performance.
\begin{table}[h]

\centering
\scriptsize

  \begin{tabular}{l | c | c | c}
    \toprule
    Methods & LLM size & Average Time Taken (s) & Number of LLM calls \\
    \midrule
    BLIP-2\cite{li2023blip} (Flan-T5-XXL)& 11B & 7.63 & 16\\
    InstructBLIP\cite{instructBLIP} (Flan-T5-XXL)& 11B & 10.01 & 16 \\
    Video-ChatGPT\cite{videochatgpt} (Vicuna-7B)& 7B & 1.87 & 1\\
    mPLUG-Owl \cite{Ye2023mPLUGOwlME} (Llama-7B)& 7B & 3.92 & 1\\
    Video-LLaVA \cite{lin2023video}(Vicuna-7B)& 7B & 2.31 & 1\\
    mPLUG-Owl+Vicuna-13B & 13B & 0.50 & 1\\    
    \midrule
    VidTFS (Vicuna-13B) & 13B & 15.17 & 4\\
    VidTFS (GPT-3.5) & Undisclosed & 6.70 & 4\\
    VidTFS (Llama3-8B) & 8B & 8.12 & 4\\
    VidTFS (Vicuna-13B using \cite{2023lmdeploy}) & 13B & 4.92 & 4\\
    VidTFS (Vicuna-13B-8bit) & 13B & 13.25 & 4\\

    \bottomrule
  \end{tabular}

\caption{Average time taken (seconds) for video inference.}
\label{tab:time_inferences}
\end{table}

\section{Qualitative Results}
\label{sec:qualitative}

\begin{table*}[tb!]

\centering
\resizebox{\textwidth}{!}{
  \begin{tabular}{p{0.3\linewidth} | p{0.7\linewidth}}
    \toprule
    Inference Task & ICL Examples  \\
    \midrule
    Goal Inference & 
    Based on the description: The person is standing on a stepladder, holding a light bulb in one hand and reaching towards the ceiling fixture with the other. There is a toolbox on the floor, and another light bulb is in his hand.
    
    Answer: 1: Replace Ceiling Light Bulb
    
    2: Replace Ceiling Fan Blades
    
    3: Install a Ceiling Medallion
    
    4: Adjust Smoke Detector
    
    5: Paint Ceiling
    
    Based on the description: The person is seated at a table covered with a large sheet of white paper. They are holding a heat gun and aiming it at a colorful arrangement of crayon pieces placed along the top edge of the paper. Then, crayon wax is melting and dripping down the paper onto a canvas below.
    
    Answer: 1: Make Melted Crayon Art
    
    2: Make Crayon Candles
    
    3: Prepare Crayon Canvas
    
    4: Make a Fresco Painting
    
    5: Paint Bookshelves\\
    \midrule
    Action Recognition & Based on the description: The human is holding a paintbrush or other painting tool, with their arm extended towards a canvas or surface, possibly leaning or sitting in front of it.
    
    Answer: 1: Painting
    
    2: Drawing
    
    3: Sketching
    
    4: Coloring
    
    5: Crafting
    
    Based on the description: The human is sitting on a bicycle, hands on the handlebars, feet on the pedals, and body leaning forward.
    
    Answer: 1: Cycling
    
    2: Biking
    
    3: Wheeling
    
    4: Pedaling
    
    5: Riding
    \\
    
    \bottomrule
  \end{tabular}
}
  
\caption{ICL examples used in open-vocabulary inference tasks}
\label{tab:icl_prompt}
\end{table*}
\begin{table*}[ht]

\centering
\resizebox{\textwidth}{!}{
  \begin{tabular}{p{0.3\linewidth} | p{0.7\linewidth}}
    \toprule
    Inference Task & Prompt  \\
    \midrule
    $\varphi_v$ or $\varphi_f$ to infer top-K hypotheses, $\mathcal{H}$ / $\mathcal{\ddot{H}}$ or final answer $h$ & I want to perform <TASK NAME> after observing some visual descriptions.

    <ICL EXAMPLE>
    
    Based on the description: < $\mathcal{D}$ or $\ddot{\mathcal{D}}$ >
    
    \{Based on these options: <$\mathcal{H}$ $\oplus$ $\ddot{\mathcal{H}}$ $\oplus$ $h_c$>\}
    
    List the most likely <K NUMBER> correct <TARGET> without any explanation. Answer: \\
    
    \midrule
    $\varphi_l$ to generate hypothesized steps, $\mathcal{S}$ & ``Briefly list down the steps to perform < $\mathcal{H}$ >.

    List down in point format without require any specific quantity or unit.'' \\
    
    \bottomrule
  \end{tabular}
  }
  
\caption{Prompt template for LLM used in both goal and action inference tasks. The placeholder <TASK NAME> also denote as $\phi$ which is replaceable with the specific inference task name (e.g. goal inference, action recognition), whereas <ICL EXAMPLE> is for insert the In-Context Learning (ICL) example when infer the hypotheses only, otherwise, it will be empty when not required. The <$\mathcal{D}$ or $\ddot{\mathcal{D}}$> indicate the input of visual descriptions. For \{Based on these options: <$\mathcal{H}$ $\oplus$ $\ddot{\mathcal{H}}$ $\oplus$ $h_c$>\}, it is only applied when there is an option list provided to prompt LLM select the final inference from the hypotheses. The <K NUMBER> is an integer value to control how many hypotheses suppose be inferred. Lastly, the <TARGET> is the term of desired outcome (e.g. ``action goal'' or ``ongoing action'') to help LLM understand the specific output for the inference task.}
\label{tab:llm_prompt8_template}
\end{table*}

\begin{figure*}[h!t]
  \centering
   \includegraphics[width=0.9\linewidth]{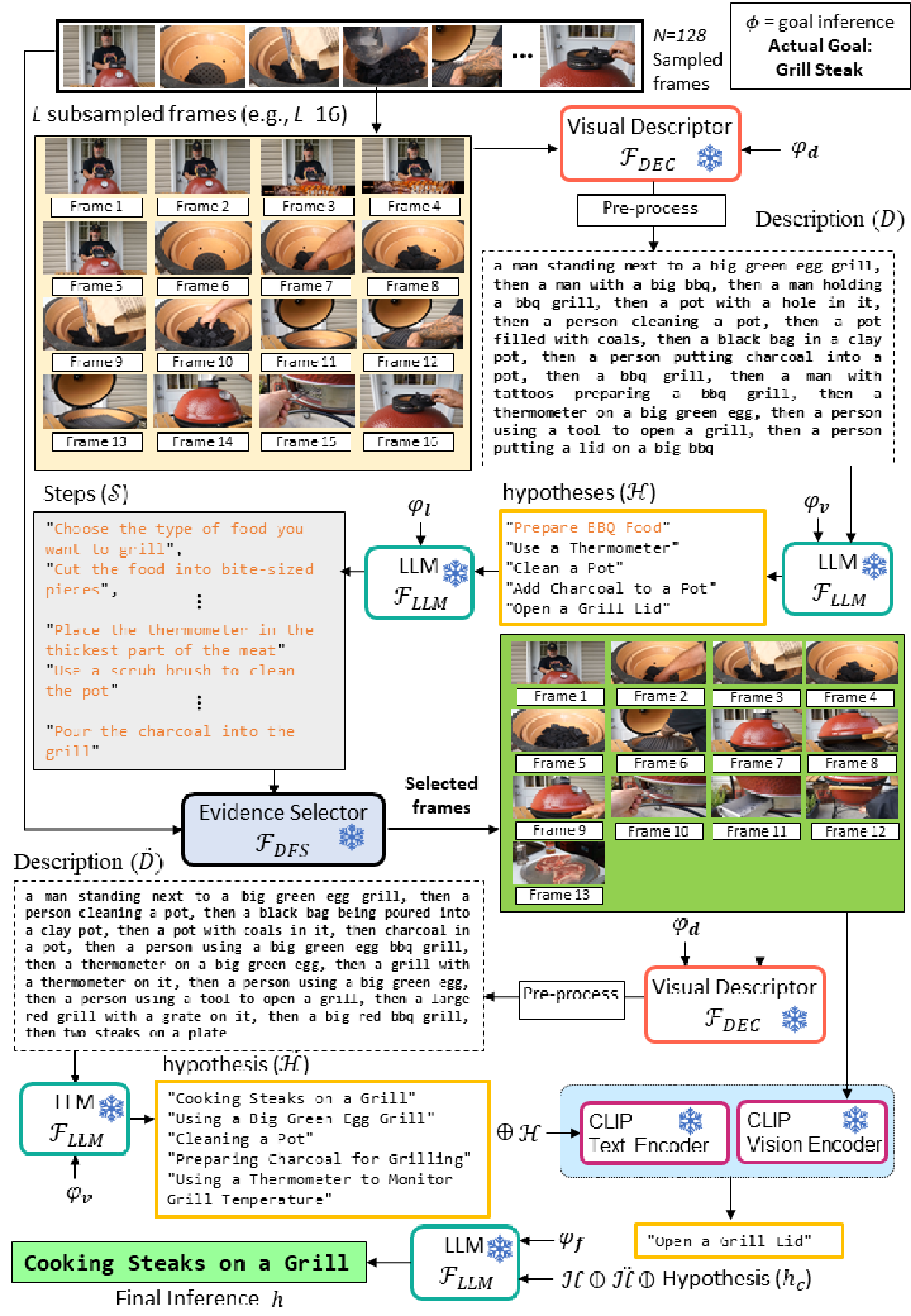}
   \caption{Qualitative example of goal inference by VidTFS (V13B) framework on CrossTask video ($\rho$ = 50\%). We demonstrate the frames selection process of the evidence selector which leads to better hypotheses and final inference: ``\textbf{Cooking Steaks on a Grill}'' \textit{vs} ground truth: ``\textbf{Grill Steak}'' (obtain 86.3 SBERT score). We can see the selected frames are more relevant to the grill with charcoal and steak after frame selection process.}
   \label{fig:dl91S-X_VJs_full_qualitative_diagrams}
\end{figure*}

\begin{figure*}[h!t]
  \centering
   \includegraphics[width=0.9\linewidth]{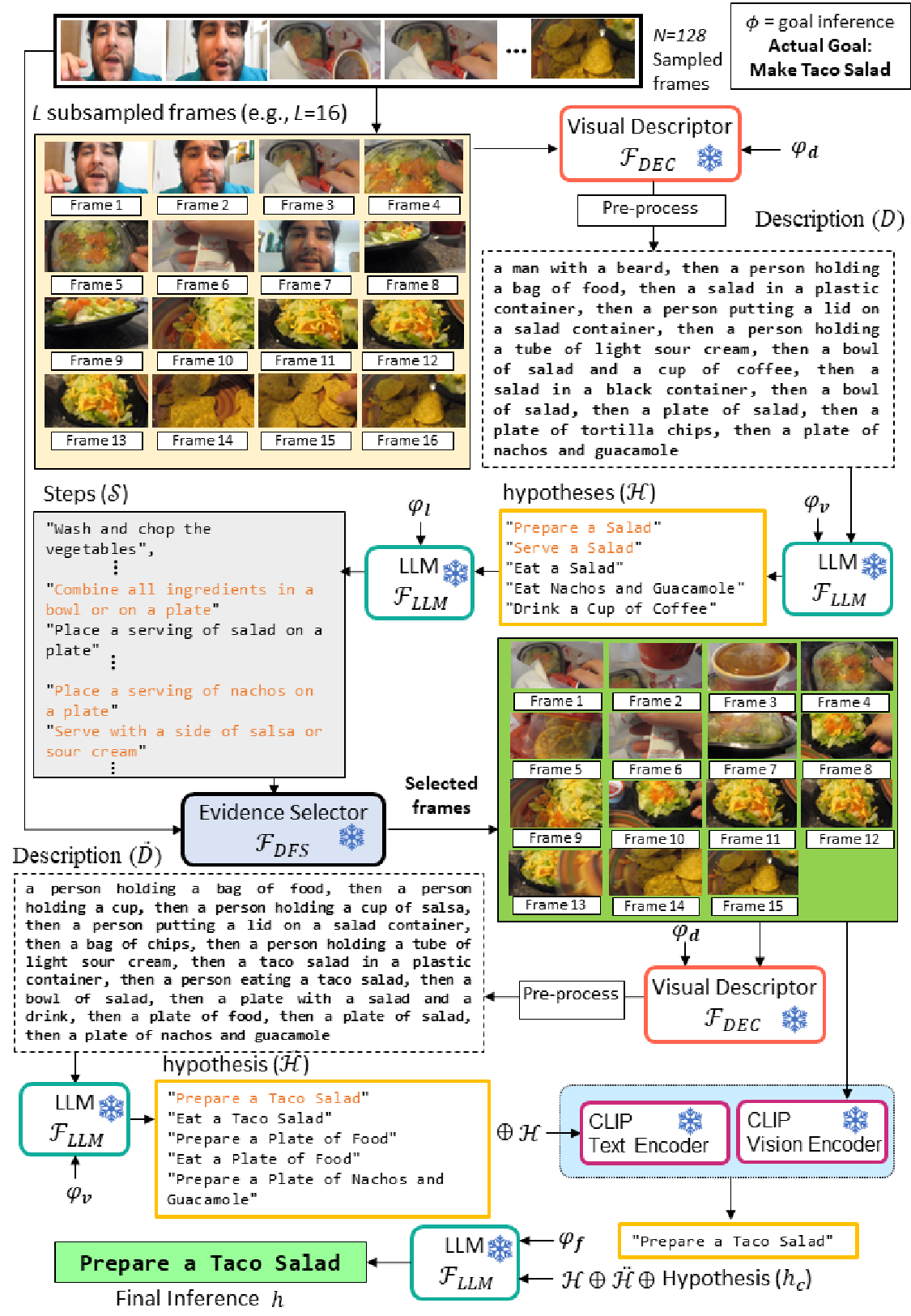}
   \caption{{Qualitative example of goal inference by VidTFS (V13B) framework on CrossTask video ($\rho$ = 50\%). We can noticed the initial sampled frames that related to a man with beard are filtered out after frame selection process as it is not relevant to the goal. We also can find the inference direction shift from salad only to taco salad related after matching the frames with the hypothesized steps that contained of taco or nachos related steps.}}
   \label{fig:k5alqWUASM8_full_qualitative_diagrams}
\end{figure*}

\begin{figure*}[h!t]
  \centering
   \includegraphics[width=0.9\linewidth]{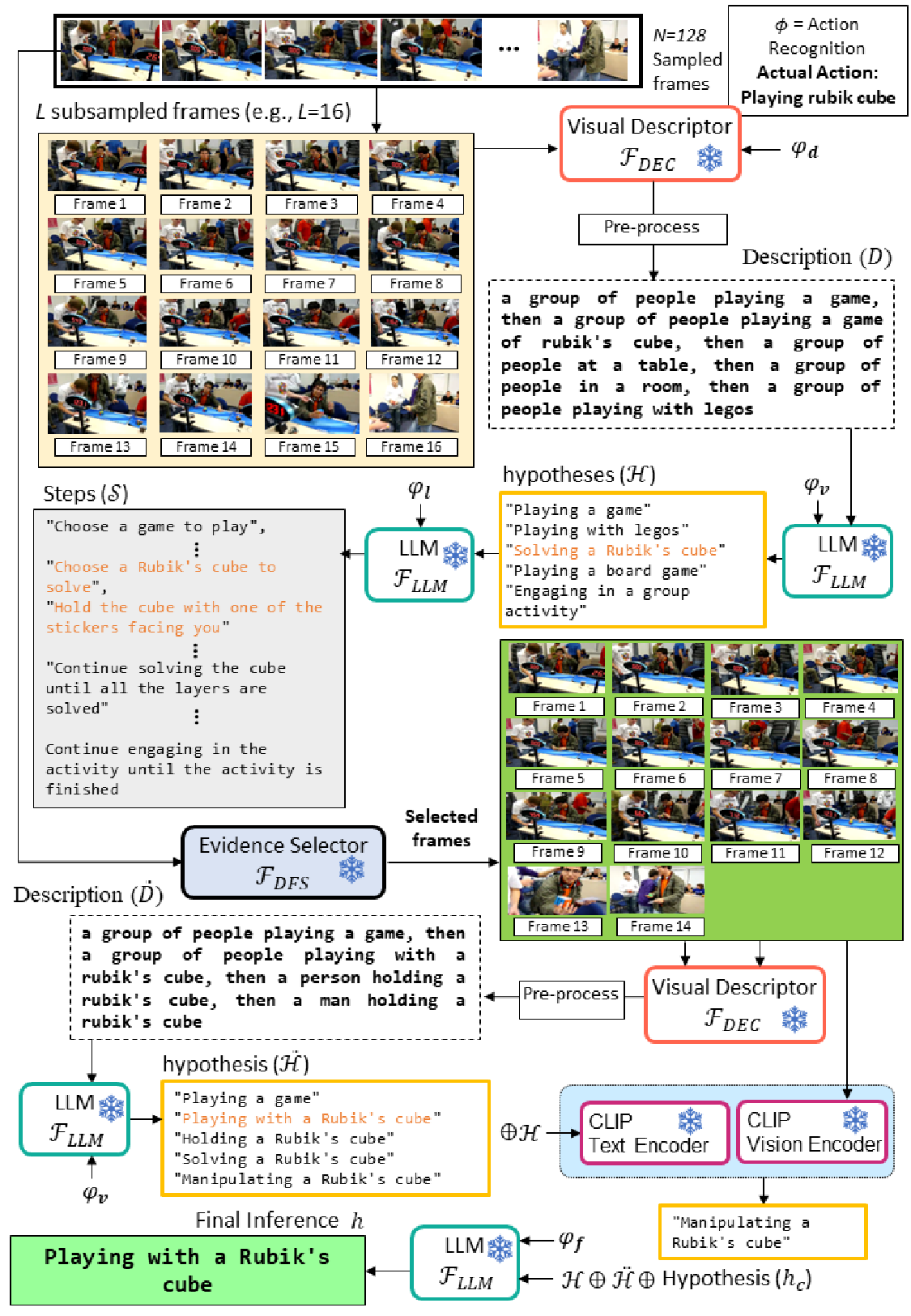}
   \caption{{Qualitative example of action recognition by VidTFS (V13B) framework on a video ($\rho$ = 100\%) from ActivityNet. Although video action recognition task is more straightforward, it is still challenging when infer on longer untrimmed video that contained many ongoing actions. We can see that initial hypotheses $\mathcal{H}$ is uncertain about the action, whereas $\mathcal{\ddot{H}}$ inference after frame selection process is more certain that the action is related to the Rubik's Cube.}}
   \label{fig:xYeqvN8cihg_full_qualitative_diagrams}
\end{figure*}

\begin{figure*}[h!t]
  \centering
   \includegraphics[width=0.9\linewidth]{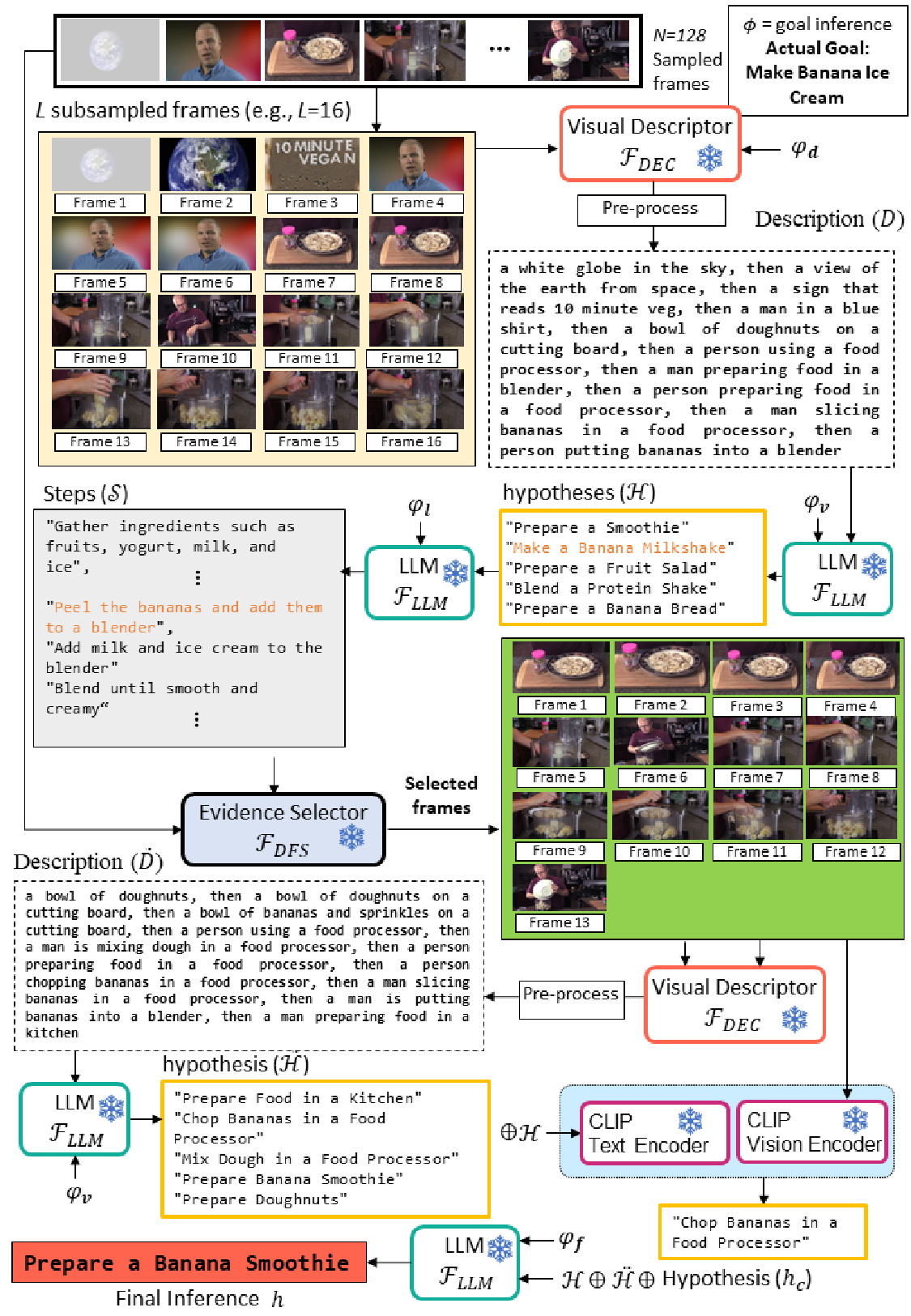}
   \caption{{Example of incorrect goal inference by VidTFS (V13B) framework on CrossTask video ($\rho$ = 30\%). We can notice that the banana slices in the bowl is wrongly recognized as ``doughnuts'' in a bowl. This suggests that a visual descriptor with better object-recognizing ability could mitigate this misidentified problem. Moreover, the ice cream related frames are not seen, the LLM is missing this important clue and hence it cannot relate to banana ice cream related goals. We also notice that the frames of \textit{"view of the earth from space"} and \textit{"a man in blue shirt"} are filtered out after frame selection process. This shows that the evidence selector is able to select the frames that are more relevant to the hypotheses.}}
   \label{fig:DZsJB5KIuZs_full_qualitative_diagrams}
\end{figure*}

We present a few more detailed qualitative examples as in Figure \ref{fig:dl91S-X_VJs_full_qualitative_diagrams}, \ref{fig:k5alqWUASM8_full_qualitative_diagrams}, 
and \ref{fig:xYeqvN8cihg_full_qualitative_diagrams} 
that included detail intermediate outputs along the inference process in the VidTFS framework. 
We also show a failure example in Figure \ref{fig:DZsJB5KIuZs_full_qualitative_diagrams}.
Best viewed on computer full screen.

\clearpage






\end{document}